\documentclass[preprint,12pt]{elsarticle}
\usepackage{pifont}  
\usepackage{float}
\newcommand{\cmark}{\ding{51}}  
\newcommand{\xmark}{\ding{55}}  

\usepackage{amssymb}
\usepackage{amsmath}

\usepackage{float}
\usepackage{booktabs}
\usepackage{tabularx}
\usepackage{multirow}
\usepackage[table,x11names]{xcolor}
\usepackage{eurosym}             
\usepackage{caption}
\usepackage[export]{adjustbox}   

\journal{Medical Image Analysis}

\begin{document}

\begin{frontmatter}



\title{Neural Implicit Heart Coordinates: 3D cardiac shape reconstruction from sparse segmentations}


\author[label1]{Marica Muffoletto \corref{cor1}}
\author[label1]{Uxio Hermida}
\author[label1]{Charlène Mauger}
\author[label1]{Avan Suinesiaputra}
\author[label1]{Yiyang Xu}
\author[label2]{Richard Burns}
\author[label3,label4]{Lisa Pankewitz}
\author[label5]{Andrew D McCulloch}
\author[label2,label6]{Steffen E Petersen}
\author[label7]{Daniel Rueckert}
\author[label1]{Alistair A Young}

\cortext[cor1]{Corresponding author.}
\affiliation[label1]{organization={School of Biomedical Engineering and Imaging Sciences}, 
            addressline={King's College London}, 
            city={London}, 
            country={UK}}
\affiliation[label2]{organization={William Harvey Research Institute, NIHR Barts Biomedical Research Centre}, 
            addressline={Queen Mary University}, 
            city={London}, 
            country={UK}}
\affiliation[label3]{organization={Simula Research Laboratory}, 
            city={Oslo}, 
            country={Norway}}
\affiliation[label4]{organization={Department of Informatics}, 
            addressline={University of Oslo}, 
            city={Oslo}, 
            country={Norway}}
\affiliation[label5]{organization={Department of Bioengineering}, 
            addressline={University of California},
            city={San Diego}, 
            country={USA}}
\affiliation[label6]{organization={Barts Heart Centre, Barts Health NHS Trust}, 
            addressline={St Bartholomew's Hospital, West Smith}, 
            city={London}, 
            country={UK}}
\affiliation[label7]{organization={Technical University of Munich}, 
            city={Munich}, 
            country={Germany}}
\begin{abstract}
Accurate reconstruction of cardiac anatomy from sparse clinical images remains a major challenge in patient-specific modeling. While neural implicit functions have previously been applied to this task, their application to mapping anatomical consistency across subjects has been limited. In this work, we introduce Neural Implicit Heart Coordinates (NIHCs), a standardized implicit coordinate system, based on universal ventricular coordinates, that provides a common anatomical reference frame for the human heart. Our method predicts NIHCs directly from a limited number of 2D segmentations (sparse acquisition) and subsequently decodes them into dense 3D segmentations and high-resolution meshes at arbitrary output resolution. Trained on a large dataset of 5,000 cardiac meshes, the model achieves high reconstruction accuracy on clinical contours, with mean Euclidean surface errors of 2.51±0.33 mm in a diseased cohort (n=4549) and 2.31±0.36 mm in a healthy cohort (n=5576). The NIHC representation enables anatomically coherent reconstruction even under severe slice sparsity and segmentation noise, faithfully recovering complex structures such as the valve planes. Compared with traditional pipelines, inference time is reduced from over 60 s to 5–15 s. These results demonstrate that NIHCs constitute a robust and efficient anatomical representation for patient-specific 3D cardiac reconstruction from minimal input data.
\end{abstract}


\end{frontmatter}



\section{Introduction}
\label{intro}

Cardiac shape reconstruction is a key area of research in medical imaging, aimed at generating accurate three-dimensional anatomical heart shapes from clinical imaging data. However, the heart’s complex structure and dynamic motion make anatomical reconstruction particularly challenging. In particular, standard cardiovascular magnetic resonance imaging is typically limited by a sparse acquisition, where only a small number of slices is acquired (typically 10 short-axis slices, spaced 10-20 mm apart, and three long-axis images in four-chamber, two-chamber, and three-chamber orientations) with misalignment between slices due to inconsistent breath-hold positions. 

Cardiac mesh models play a central role in reconstructing heart geometry and modeling disease-related structural changes (i.e., remodeling). These mesh models represent both the surfaces and volumes of the cardiac chambers and are essential to the development of personalized medicine through digital twins in order to analyze heart function, simulate blood flow dynamics, compute electrophysiological activation properties and biomechanical analyses of stress and energy consumption \cite{corral2020digital}. Accurate cardiac shape models also enable improved quantification of structural abnormalities, and enhance the prediction of future adverse cardiac events \cite{mauger2022multi}. In computational physiology applications, mesh models are obtained by applying meshing algorithms to segmented medical images \cite{strocchi2020publicly}. Since the resulting meshes can vary substantially between patients, in both geometry and number of vertices and their connectivity, several groups have proposed a Universal Ventricular Coordinate system (UVC) enabling consistent mapping of physiological information such as fiber orientation, mechanical properties, scar regions, and activation times between geometries \cite{bayer2018universal,schuler2021cobiveco,pankewitz2024universal}. Given a mesh, UVCs can be calculated by solving partial differential equations such as the Eikonal equation to give a consistent anatomical coordinate system for personalized cardiac digital twin analysis. However, the process of computing UVCs currently depends on having a 3D high resolution mesh as input.

In medical imaging, Statistical Shape Models (SSMs) have traditionally been used to reconstruct heart shapes and capture their statistical variations \cite{lotjonen2004statistical, mauger2019right, bai2015bi}. These approaches rely on a consistent mesh topology—meaning a standardized mapping between edges and vertices that remains the same across all meshes—which is then personalized for each case. Such templates were typically constructed using non-rigid registration or point-to-surface distance minimization techniques. 

Inspired by advances in computer vision, where traditional mesh reconstruction methods based on dense, noise-sensitive point clouds \cite{carr2001reconstruction, ummenhofer2015global, kazhdan2013screened} have been increasingly replaced by deep learning approaches, the medical imaging field has begun to adopt similar strategies. Convolutional neural networks (CNNs) and graph-based or geometric deep learning approaches have become popular for 3D reconstruction tasks. For example, CNNs have been used to predict volumetric representations or deformation fields on mesh templates, while graph convolutional networks (GCNs) have been applied to reconstruct anatomical structures from mesh vertices \cite{kong2021deep}. In cardiac imaging, approaches include grid-based 3D U-Nets for ventricular mesh reconstruction \cite{xu2019ventricle}, Point2Mesh-style CNNs for deforming meshes from sparse point clouds \cite{beetz2022point2mesh, hanocka2020point2mesh}, and variational mesh autoencoders for reconstructing cine MR images \cite{beetz2022interpretable}. These approaches often leverage databases of consistently shaped meshes across multiple patients, full dense data at training time, and require the same set of slices at training time and test time, limiting their use in scenarios with noisy input data.

Neural implicit functions (NIFs) provide an alternative by representing shapes as continuous functions over 3D space, typically parametrized by multi-layer perceptrons (MLPs). This continuous representation enables querying at arbitrary resolution and location, making them particularly robust to sparse, partial, or non-uniform input data \cite{chen2019learning, mescheder2019occupancy, hanocka2020point2mesh}. In computer vision, NIFs have been successfully applied to reconstruct objects from incomplete point clouds, multi-view images, or depth maps, often achieving higher fidelity and smoother surfaces compared to voxel or mesh-based approaches. Early methods trained separate networks per shape \cite{Huang2023, alblas2023implicit}, while more recent approaches incorporate shape priors via latent codes that capture shape-specific geometry across multiple objects \cite{amiranashvili2022learning, park2019deepsdf, mescheder2019occupancy, chen2019learning, amiranashvili2024learning}. 
Applying NIFs in medical imaging, however, poses several distinct challenges. These include the heterogeneity of acquisition protocols (e.g., variations in slice spacing, orientation, and alignment), severe anisotropy, motion artifacts, limited sample sizes, the need for robust regularization to prevent overfitting to the shape prior, and difficulties in meaningful quantitative evaluation when dense ground-truth volumes are unavailable. Nonetheless, NIFs have been successfully applied in this domain, for instance in brain segmentation \cite{vyas2025fit}, cardiac label map reconstruction from just a few CMR slices \cite{muffoletto2023neural}, and in jointly reconstructing MR images and segmentations \cite{stolt2023nisf}.

Compared to graph-based architectures, which operate on discrete mesh vertices with predefined connectivity, MLP-based NIFs learn a continuous, resolution-agnostic function over the entire 3D domain. This enables flexible surface representation, arbitrary query points, and robust handling of sparse or non-uniformly sampled clinical data \cite{Hu2022, amiranashvili2022learning}. Such properties are particularly advantageous in clinical scenarios with limited slices, varying orientations, or patient-specific anatomical variations. While GNNs excel at modeling local mesh deformations, they are less flexible for arbitrary-resolution reconstruction and may struggle when inputs are sparse or partially acquired. Therefore, MLP-based NIFs can provide a better solution to achieve resolution independence, and robustness for medical shape reconstruction.

\begin{figure}[H]
    \centering
    \includegraphics[width=\textwidth]{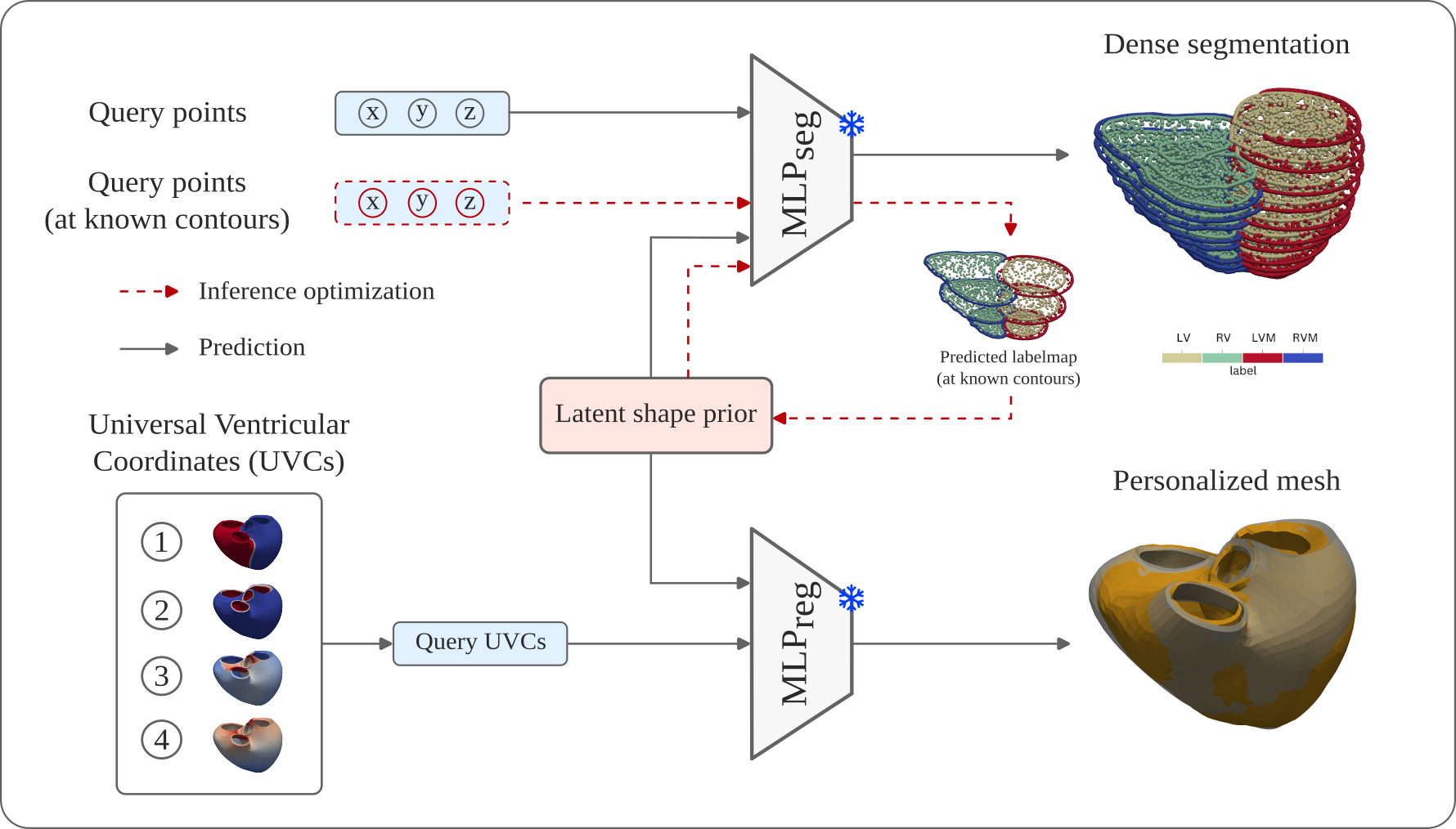}
    \caption[Overview]{Overview of our pipeline for simultaneous prediction of dense segmentations and 3D personalized meshes from sparse segmentations. During inference, first, coordinates from known points (e.g., segmentation contours) are used together with a previously trained MLP model (MLP$_{seg}$) to predict occupancy at those points and optimize a latent shape prior vector (see red dotted arrows). This process is referred to as inference optimization. Once the optimal latent shape vector has been found, a second MLP (MLP$_{reg}$) is used together with queried Universal Ventricular Coordinates (UVCs) to predict a 3D personalized mesh. The optimal latent shape prior vector can also be used by the MLP$_{seg}$ to generate a dense segmentation. As both MLPs learn a continuous mapping from their respective inputs to outputs, both dense segmentation maps and personalized 3D models can be generated at any arbitrary resolution.}
    \label{fig:1A}
\end{figure}

In this work, we propose a novel method for the reconstruction of 3D cardiac anatomy and mesh models from sparse CMR image data, using NIFs to predict both geometry (i.e., dense segmentation) and shape (i.e., mesh). The method has the advantage of only requiring segmentation data from any acquired slices in any orientation. Beyond reconstructing anatomical surfaces, our approach also enables the computation of UVCs directly from the reconstructed implicit representation. This is made possible because the continuous nature of the NIF allows for the generation of complete, topologically consistent meshes, from which coordinate systems can be derived. To our knowledge, this is the first method capable of computing UVC maps from sparse or partial segmentation data, without the need for a pre-existing 3D mesh as input. This work extends our previous study \cite{muffoletto2023neural} in three key ways:

\begin{itemize}
    \item We introduce a regression module (MLP${reg}$) that learns the mapping from UVCs to Cartesian coordinates. This module is integrated with our previously proposed segmentation network (MLP${seg}$), which predicts anatomical labels from spatial coordinates. As shown in Fig.\ref{fig:1A}, the two MLPs are jointly trained within a unified pipeline that leverages a shared latent shape prior, enabling simultaneous reconstruction of dense label maps and personalized 3D meshes from sparse segmentations.
    \item We introduce Neural Implicit Heart Coordinates (NIHCs). NIHCs, like UVCs, provide a standardized reference system across predicted heart geometries, enabling simple estimation directly from sparse segmentations. 
    \item We scale to a large cohort (from UK Biobank \cite{petersen2016uk,raisi2021cardiovascular}) of 10,000 cases, including both healthy and diseased cases, by sampling points from real clinical contours, allowing for more flexible and robust shape reconstruction.
\end{itemize}

\section{Method} \label{method}

\subsection{Data}

An overview of the data generation pipeline is shown in Fig.\ref{fig:4B}. Briefly, from a collection of UK Biobank CMR images, contours of the left and right ventricle were extracted, corrected for breath-hold misalignment, and used to fit a template mesh to each case. The subsequent meshes were then used to calculate a mean population shape, for which the UVCs were calculated using the CobivecoX methodology \cite{pankewitz2024universal}. The resulting UVCs were then mapped to each case mesh, and a set of synthetic contours were generated to be used during training and testing.

The UK Biobank imaging study includes over 100,000 participants aged 40-69 years, recruited in the UK between 2006 and 2010 \cite{raisi2021cardiovascular, petersen2016uk}. We used images under access approval number 2964. These were acquired using a 1.5 Tesla scanner (MAGNETOM Aera, Syngo Platform VD13A, Siemens Healthineers AG, Erlangen, Germany) with retrospective electrocardiogram (ECG) gating. Short-axis (SAX) cine images were obtained in a contiguous stack with an 8 mm slice thickness and 10 mm spacing, covering the left ventricle (LV) and right ventricle (RV). Additionally, three long-axis (LAX) cine slices were acquired with a 6 mm slice thickness, oriented in the four-chamber (4CH), two-chamber (2CH), and three-chamber (3CH) views - see Fig.\ref{fig:4B}-1. In this work, 45,683 participants with imaging studies were available.

Contours and landmarks were automatically identified using the cvi42 post-processing software (Version 5.11 1505, Circle Cardiovascular Imaging Inc., Calgary, Canada) - see Fig.\ref{fig:4B}-2. This software employs deep learning convolutional neural network algorithms for fully automated analysis, which has been previously validated \cite{bottcher2020fully}. Contours were identified in both short- and long-axis images, including two-chamber, three-chamber, and four-chamber views. Breath-hold misalignment between slices was automatically corrected using a previously proposed correction method \cite{sinclair2017fully}. The misalignment correction method was limited to in-plane shifts of the segmentation masks in each slice to align with the mask of other slices in 3D space - see Fig.\ref{fig:4B}-3.
Landmarks for the tricuspid and aortic valves were defined on the four-chamber and three-chamber LAX views, respectively, while mitral valve points were identified on all LAX images. An LV endocardial apex point was determined on the four-chamber image. Since the machine learning algorithms did not detect the RV free wall myocardium and this is instead required for the computation of the UVCs, the RV free-wall epicardial surface was approximated by radially displacing the RV endocardial points by 3 mm, corresponding to the average normal RV myocardial thickness in adults \cite{foale1986echocardiographic}. Ventricular volumes and LV mass were computed by cvi42 using short-axis slice summation. The end-diastolic (ED) and end-systolic (ES) frames were identified as those with the highest and lowest reported volumes.

Biventricular shape models at ED were estimated for 41,659 Biobank cases with valid cvi42 data using the pipeline described previously in \cite{burns2024genetic}. Briefly, a template shape model consisting of a subdivision surface mesh, was automatically customized for each case following the iterative mesh fitting approach described in \cite{mauger2018iterative}, which uses a diffeomorphic least squares minimization of the distances between a template shape model and the contour points from all short-axis and long-axis slices. The template mesh had 5,810 vertices describing both ventricles. Residual misalignments and segmentation errors (leading to inconsistencies between short and long axis segmentation masks) resulted in a mean Euclidean distance between the segmented contours and the optimized mesh surfaces of 2.2mm on average - see Fig.\ref{fig:4B}-4. Each mesh was aligned by transforming its coordinates into a common cardiac coordinate space using an affine transformation. A mean shape was then obtained - see Fig.\ref{fig:4B}-5. For more details, see Supplementary Materials.

We selected all participants with prevalent cardiovascular disease (CVD) according to hospital episode statistics to be part of the evaluation test dataset. Based on codes ICD10, ICD9, and OPCS4, we included cases of heart failure, myocardial infarction or ischaemic disease, ventricular arrhythmia, conduction defects, atrial fibrillation, and diabetes mellitus. This resulted in 4,549 cases with prevalent disease. In addition, we also randomly selected 5,576 cases with no record of prevalent cardiovascular disease (No-CVD) as a control subgroup of the evaluation test dataset. This test set corresponds to the same one used in our previous work \cite{muffoletto2024evaluation}, to allow for direct comparison. For the training dataset, we randomly selected 5,000 cases (with no record of cardiovascular disease but separate from the control evaluation sub-group). In total, out of the original 45,683 cases available, we used 15,125 cases: 5,000 for training with a mean $\pm$ std. LV volume of 146.5 $\pm$ 30.1 mL, RV volume of 147.4 $\pm$ 32.0 mL and LV mass 111.2 $\pm$ 24.25 g; 4,549 diseased test cases with LV volume 155.5 $\pm$ 35.5 mL, RV volume 153.3 $\pm$ 34.6 mL, LV mass 122.9 $\pm$ 26.2 g; 5,576 healthy test cases with LV volume 155.9 $\pm$ 50.6 mL, RV volume 157.5 $\pm$ 54.2 mL and LV mass 118.6 $\pm$ 37.1 g. 

\subsection{Estimation of Neural Ventricular Coordinates}

To map each shape in the dataset to a common reference system, we use UVCs, a framework originally introduced by Bayer et al. \cite{bayer2018universal} for biventricular meshes. This method defines four coordinates - transventricular, transmural, roational, and apicobasal - each derived from solutions to Laplace’s equation.  UVCs are designed to provide a standardized coordinate system with a unique set of coordinates for all points in the heart, which is able to be mapped onto any heart geometry whilst maintaining bijectivity, continuity, normalization, linearity, and consistency. Geodesic distances are defined to be proportional to changes in coordinate values, and anatomical landmarks are consistently mapped across different heart shapes, making the system robust to variations in morphology. In our work, we follow the definition of UVC maps given by Pankewitz et al. \cite{pankewitz2024universal, schuler2021cobiveco} (CobivecoX method), which addresses inconsistencies in the original formulation and extends to the outflow tracts, valve annuli and intervalvular anatomy. 

We computed CobivecoX UVC maps for the mean Biobank mesh, using the method described in  \cite{pankewitz2024universal} - see Fig.\ref{fig:4B}-6 (for further details, see Supplementary Material Fig.~\ref{fig:app-uvc}). Briefly, UVC 1 is a transventricular binary coordinate set to 0 for the LV and 1 for the RV. The transition occurs at the center of the septum, in order to ensure symmetry in the transmural, rotational, and apicobasal coordinates across both ventricles. UVC 2 is a transmural coordinate defined to be 0 on the epicardial surface and 1 on the endocardial surface of both ventricles. In the septum, UVC 2 increases from 0 at the center of the septum, so that both sides of the septal endocardium share the same value of 1. In the ventricles, UVC 3 is a rotational coordinate which counter-rotates in the LV and RV free walls, ranging from 0 at the posterior junction of the LV and RV, to 2/3 at the anterior junction. In the septum, UVC 3 ranged from 2/3 in the anterior to posterior direction. At the base, UVC 3 defines an intervalvular inflow-outflow coordinate ranging from 1 at the outflow annulus, to 1.5 at the inflow annulus. UVC 4 defines the apex to base coordinate in the ventricles, ranging from 0 at the apex to 1 at the base. In the intervalvular region, UVC 4 defines a free-wall to valve coordinate ranging from 1 at the surface nearer the free wall to 1.5 at the surface nearer the septum.

\begin{figure}[H]
    \centering
    \includegraphics[width=\textwidth]{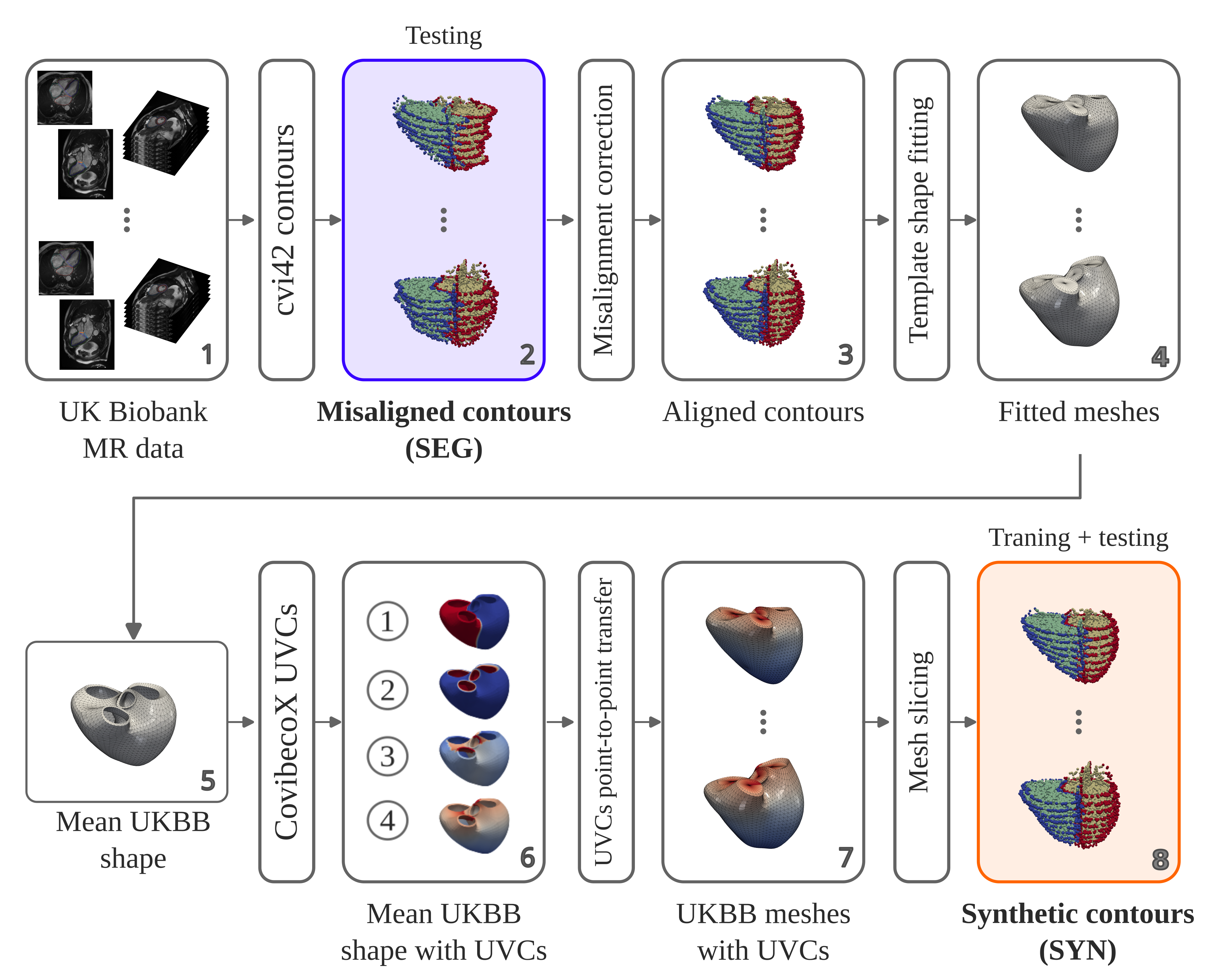}
    \caption[Data preparation]{Data generation pipeline. First, a dataset of image-derived contours (referred to as \textbf{SEG} dataset) was created by extracting contours (using Circle cvi42 postprocessing software Version 5.11 1505). This dataset was used during testing. Second, a mesh-derived dataset (referred to as \textbf{SYN}) was generated by fitting a template mesh to each case via a diffeomorphic fitting method \cite{mauger2018iterative}, after having corrected for breath-hold misalignment following ~\cite{sinclair2017fully}. After alignment, an average UKBB shape was computed, and Universal Ventricular Coordinates (UVCs) were calculated using the CovibecoX method \cite{pankewitz2024universal} and transferred to each mesh using point-to-point correspondences. The final SYN dataset consists of synthetic contours generated by slicing these meshes and assigning labels based on the UVCs. The SYN dataset was used during both training and testing.}
    \label{fig:4B}
\end{figure}

Due to the point-to-point correspondence, UVC coordinates from the average UK Biobank mesh were mapped to each patient-specific case, defining the Neural Ventricular Coordinates used to train our NIF network - see Fig.\ref{fig:4B}-7.

\subsection{Method for combined segmentation and regression}

Our method integrates two primary tasks: classification and regression. The classification component (SEG-NIF) builds upon the architecture proposed by Amiranashvili et al. \cite{amiranashvili2024learning} and its application to cardiac shape modeling from our previous work \cite{muffoletto2023neural}. The regression component (UVC-NIF) and its integration with SEG-NIF represent a novel contribution of this study.
\begin{enumerate} 
\item SEG-NIF uses a multilayer perceptron ($MLP_{\text{seg}}$) that outputs occupancy probabilities for biventricular semantic segmentation labels, taking as input Cartesian cardiac coordinates along with a latent vector $h$. The segmentation labels used in our study were: background (BG), left ventricle blood pool (LV), right ventricle blood pool (RV), left ventricle myocardium (LVM), and right ventricle myocardium (RVM). These labels were derived based on the transventricular UVC-1 coordinate mapped onto the reference mesh models: points inside the endocardial surface were labeled as LV or RV, points outside the epicardial surface were assigned the BG label, and points in between were labeled as LVM or RVM. 
\item UVC-NIF uses a second multilayer perceptron ($MLP_{\text{reg}}$), which takes as input UVC coordinates together with the shared latent code and predicts the corresponding Cartesian cardiac coordinates. This forms the regression task. \end{enumerate} 
Both $MLP_{\text{seg}}$ and $MLP_{\text{reg}}$ consist of eight residual layers, each with 128 hidden units. The former outputs 5 channels corresponding to the segmentation labels, while the latter outputs 3 channels corresponding to the global coordinates.

\begin{figure} [H]
    \centering
    \includegraphics[width=\textwidth]{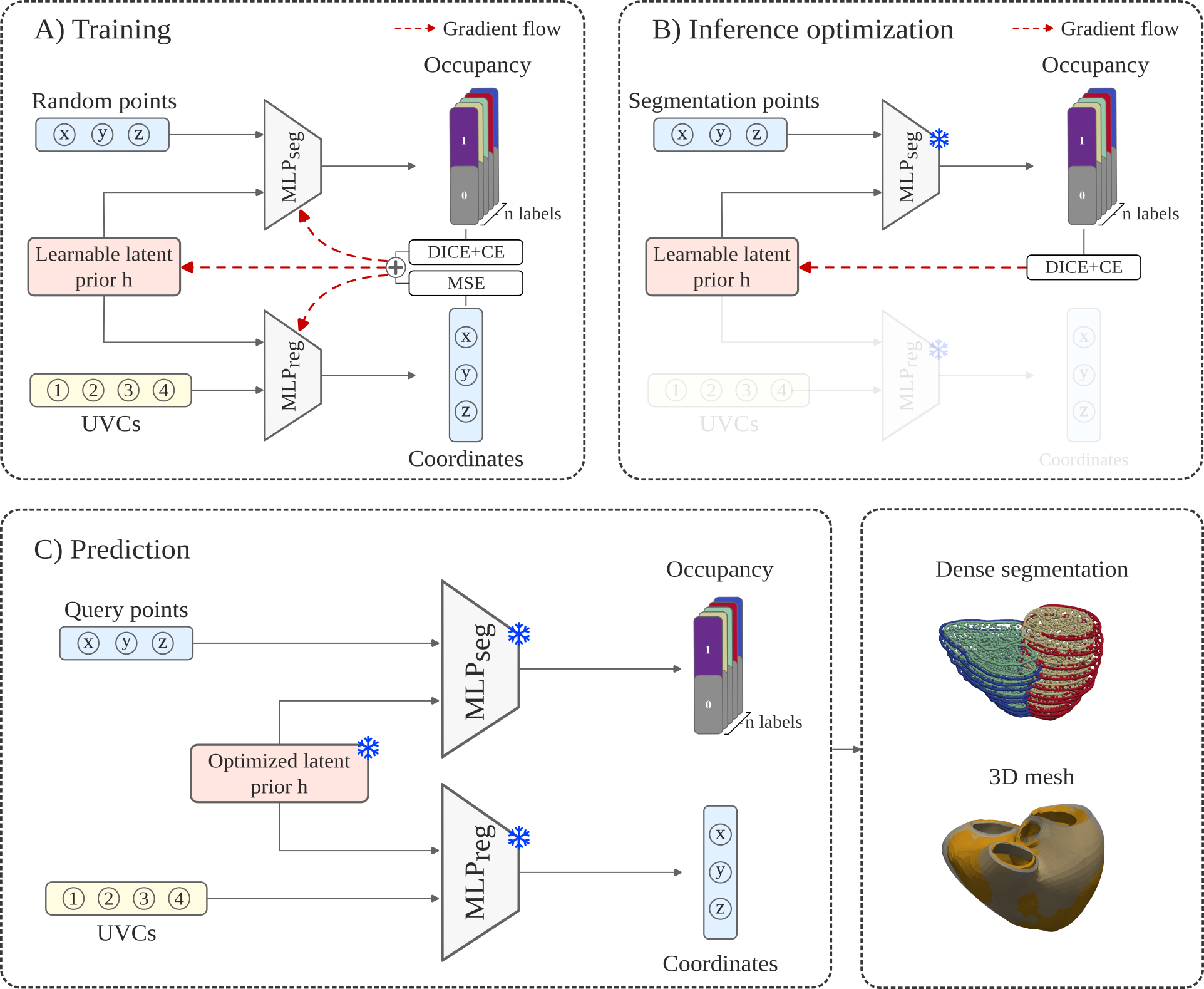}
    \caption[Methodology]{Overview of the proposed pipeline. During training, $MLP_{\text{seg}}$ and $MLP_{\text{reg}}$ are jointly optimized using a combined loss for the segmentation and regression tasks. Backpropagation updates the weights of both MLPs as well as the learnable latent prior $h$. During inference optimization, only $MLP_{\text{seg}}$ is used (kept frozen) to refine the latent prior based on the segmentation task. Finally, during prediction, both MLPs are frozen, and query points (in Cartesian or UVC coordinates) are used to reconstruct either a label map or a 3D mesh.}
    \label{fig:4C}
\end{figure}

Fig.~\ref{fig:4C} illustrates the overall procedure of our method, which is divided into three stages: training, inference optimization, and prediction. During training, the network learns a shared (classification/regression) shape prior $h$ per subject. In the inference optimization stage, the latent code is adapted to a new subject by optimizing $h$ based on the available input data. Finally, at prediction time, the optimized latent code is combined with spatial coordinates or UVCs to reconstruct either label maps or 3D meshes.

\subsubsection{Training}

SEG-NIF was trained on approximately 55K points, consisting of mesh vertices and additional random points sampled within a bounding cube surrounding the cardiac shape. In contrast, UVC-NIF was trained on approximately 10K points, including mesh vertices and points generated between the endocardial and epicardial surfaces by selecting locations where the transmural UVC-2 coordinate falls between 0 and 1 (i.e., within the myocardium). UVC values for these intermediate points were computed by linear interpolation between the nearest endocardial and epicardial vertices. The dataset consisted of 5000 non-CVD cases, split randomly into 80\% for training and 20\% for validation.

During training (Fig.~\ref{fig:4C}-A), the models were optimized jointly to learn a shared latent code for each shape by minimizing a total loss composed of three terms:

\begin{equation}
    \mathcal{L} = \frac{1}{\lambda_{\text{seg}}} \mathcal{L}_{\text{SEG}} + \frac{1}{\lambda_{\text{reg}}} \mathcal{L}_{\text{REG}} + \lambda_{\text{prior}} \mathcal{L}_{\text{prior}}
\end{equation}

where:
\begin{itemize}
    \item as suggested by \cite{amiranashvili2024learning} and \cite{isensee2021nnu}, \(\mathcal{L}_{\text{SEG}}\) is a combined Dice and binary cross-entropy (BCE) loss, 
    \item \(\mathcal{L}_{\text{REG}}\) is a mean squared error (MSE) loss for the regression task,
    \item \(\mathcal{L}_{\text{prior}}\) is a latent prior regularization term encouraging the latent codes to follow a standard Gaussian distribution \(\mathcal{N}(0, I)\).
\end{itemize}

The latent prior regularization used during training is defined as:
\begin{equation}
    \mathcal{L}_{\text{prior}} = \frac{1}{B} \sum_{i=1}^B \| h_i \|_2^2
\end{equation}

where \(h_i\) is the latent vector associated with sample \(i\) and \(B\) is the batch size. The regularization strength \(\lambda_{\text{prior}}\) is progressively increased during the first 100 epochs according to:

\begin{equation}
    \lambda_{\text{prior}} = \min\left(1, \frac{\text{epoch}}{100}\right) \lambda_{\text{prior,max}}
\end{equation}

where \(\lambda_{\text{prior,max}}\) denotes the maximum regularization strength (1.0e-4). 

The scaling factors were set to \(\lambda_{\text{seg}} = 1.0\) and \(\lambda_{\text{reg}} = 1000.0\) to balance the contributions from the segmentation and regression tasks, respectively.

\subsubsection{Inference Optimization}

Inference optimization (Fig.~\ref{fig:4C}-B) is performed solely on the segmentation task. Semantic segmentation labels, typically obtained from MRI slices such as SAX, 4CH, 2CH, and 3CH LAX views, are used as inputs for $MLP_{\text{seg}}$. The latent vector \( h \) is then optimized via backpropagation through the SEG-NIF network to minimize the segmentation loss \(\mathcal{L}_{\text{SEG}}\).

The total loss function during this stage is: 

\begin{equation}
\label{eq:4}
    \mathcal{L} = \lambda_{\text{r}} \mathcal{L}_{\text{r}} + \lambda_{\text{BCE}} \mathcal{L}_{\text{BCE}} + \lambda_{\text{Dice}} \mathcal{L}_{\text{Dice}}
\end{equation}

where \( \mathcal{L}_{\text{r}} \) is a latent-space regularization term \(\mathcal{N}(0, I)\) - this follows a Mahalanobis formulation to better reflect the empirical latent distribution learned during training. It is defined as: 

\begin{equation}
    \mathcal{L}_{\text{reg}} = (z - \mu)^T \Sigma^{-1} (z - \mu)
\end{equation}

where \( \mu \) is the mean of the latent distribution, and \( \Sigma \) is the covariance matrix. This formulation better accounts for correlations between dimensions in the latent space and ensures that the optimized latent code remains consistent with the distribution observed in the training set.

Unlike the training-time regularizer—which follows the L2 prior proposed by \cite{park2019deepsdf} with a coefficient of \( 10^{-4} \)—the inference stage uses the Mahalanobis distance described above. This choice provides a more faithful constraint on the latent space during optimization, and we empirically set its weight to \( 10^{-2} \).

\subsubsection{Prediction}

Once the optimal latent vector $h$ for a given case is determined through inference optimization, both MLPs are used during the prediction step to predict either the labels or the mesh vertex coordinates of the cardiac surface (Fig.~\ref{fig:4C}-C). Importantly, at inference time the model does not rely on case-specific UVCs. Instead, it uses a fixed set of UVC coordinates associated with the 5,806 vertices of the template mesh. During the prediction stage, these predefined UVC coordinates (and the optimized latent code for the case) are fed to the MLPs to produce the final vertex positions, ensuring mesh-free inference, as no mesh-derived coordinates from the test case are required.

As a result, the model predicts labels for points on the epicardial or endocardial surfaces (LVM/RVM) as well as the global coordinates of the mesh vertices.

\subsection{Experiments and Metrics}

We evaluated the method using three experiments, each leveraging an evaluation dataset consisting of 10,052 cases (4,549 CVD and 5,576 No-CVD). In the first experiment, we assessed the network’s ability to reconstruct 3D geometries from sparse CMR image segmentations. The task was to determine if the network could generate heart anatomies that align with the sparse segmentation information from the image slices. Starting with segmentation masks (SEG) generated by cvi42, the inference optimization step determined the latent code \( h \) for each case by minimizing the segmentation loss using the SEG-NIF classification network. During the prediction step, the optimized latent code for each shape and the UVC coordinates at the reference mesh vertices were passed to the UVC-NIF regression network to predict a 3D mesh. We then evaluated the distance between the SEG boundaries and the predicted mesh surfaces using Euclidean distance, root mean squared error (RMSE), and symmetric/asymmetric Chamfer distance. The latter is formulated as in \cite{beetz2023multi}. These metrics were compared to those obtained from the reference meshes (REF), shape models fitted to the SEG boundaries using diffeomorphic least squares optimization. Since the reference meshes were optimized using regularized least squares, we expected the error between the reference mesh surfaces and the segmentation boundaries to be similar to that between the predicted mesh surfaces and the segmentation boundaries.

In the second experiment, we assessed the network's ability to reconstruct the reference mesh. The objective was to determine whether the inference optimization step, using sparse slices, could successfully reproduce the original geometry and mesh. As in the first experiment, segmentation masks from cvi42 were used for inference optimization, and UVC vertex coordinates were employed during the prediction step to generate the mesh. We compared the predicted mesh vertices with the reference mesh vertices using Euclidean distance, root mean squared error (RMSE), and Chamfer distance metrics. Additionally, Dice scores were computed from the labels predicted by the SEG-NIF network at the query points (mesh vertices). This point-based Dice is mathematically equivalent to the F1-score used in classical classification tasks. While it differs from the conventional voxel- or image-based Dice metric, it provides a meaningful and interpretable measure of per-point label agreement before the spatial reconstruction obtained with UVC-NIF.

To isolate errors arising from inconsistencies in the segmentation masks (such as misalignment between short- and long-axis slices or discrepancies from the reference mesh), we also generated synthetic segmentation masks (denoted as SYN) directly from the reference meshes by sampling points from the actual image slices. Opposed to the SEG data, these synthetic masks were free from misalignment. 

Finally, the third experiment involved an ablation study to evaluate model performance using only partial data from the available slices at test time. We tested subsets of slices, including LAX 3CH, LAX 4CH, and a variable number of SAX slices, to assess the impact of different input configurations on the final predictions.

During inference optimisation, the weights \( \lambda_{\text{BCE}} \) and \( \lambda_{\text{DICE}} \) were introduced to address a domain gap observed between the aligned training data (SEG) and the real misaligned ones (SYN). Hence, to balance the loss terms in Eq.~\ref{eq:4}, we empirically set \( \lambda_{\text{BCE}} = 10 \) and \( \lambda_{\text{Dice}} = 1 \)in the SYN case, and both weights to 1 in the SEG case. This is valid for all the experiments.

\section{Results}

The training took 12 hours, for 1000 iterations, on our TITAN RTX (24 GB) machine. Inference optimization typically took 1.5-5 seconds for the SYN group, 5-10s for the SEG, and prediction of mesh geometry took less than 2 seconds (on a RTX 4090 NVIDIA GPU). Compared with the time required to fit the model with the existing diffeomorphic least squares optimization (60 sec). 

\begin{figure}[H]
    \centering
    \includegraphics[width=\textwidth]{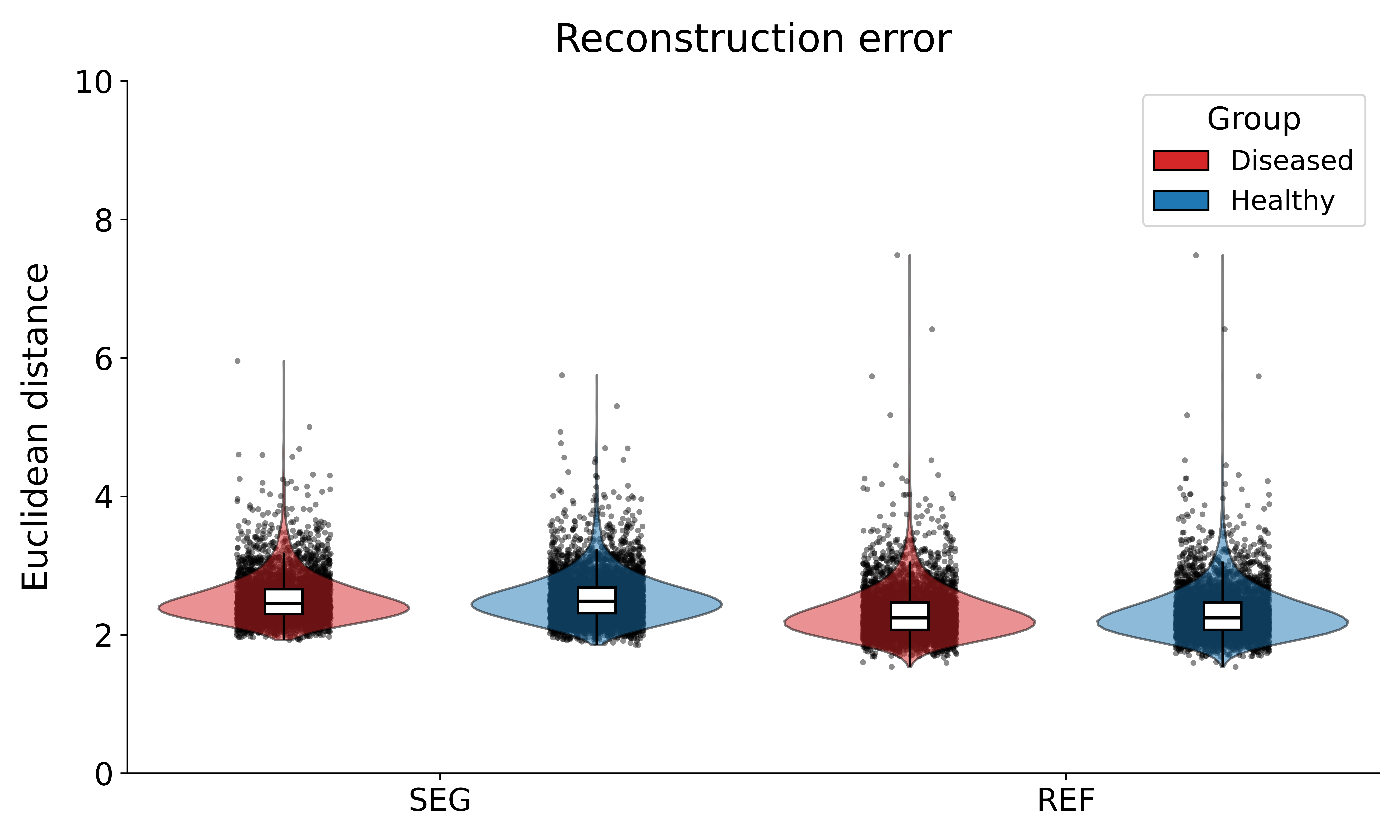}
    \caption{Plot of the Euclidean distance calculated from the reference (REF) and segmentation (SEG) meshes to the SEG contours obtained from the CMR slices.}
    \label{fig:res1}
\end{figure}

For the first experiment, the distances between the original contour points obtained from cvi42 and the predicted mesh surfaces (SEG), as well as the reference meshes (REF), are reported in Table~\ref{tab:res1}. The distribution of the ED values is illustrated in Fig.~\ref{fig:res1}.

\begin{figure}[H]
    \centering
    \includegraphics[width=\textwidth]{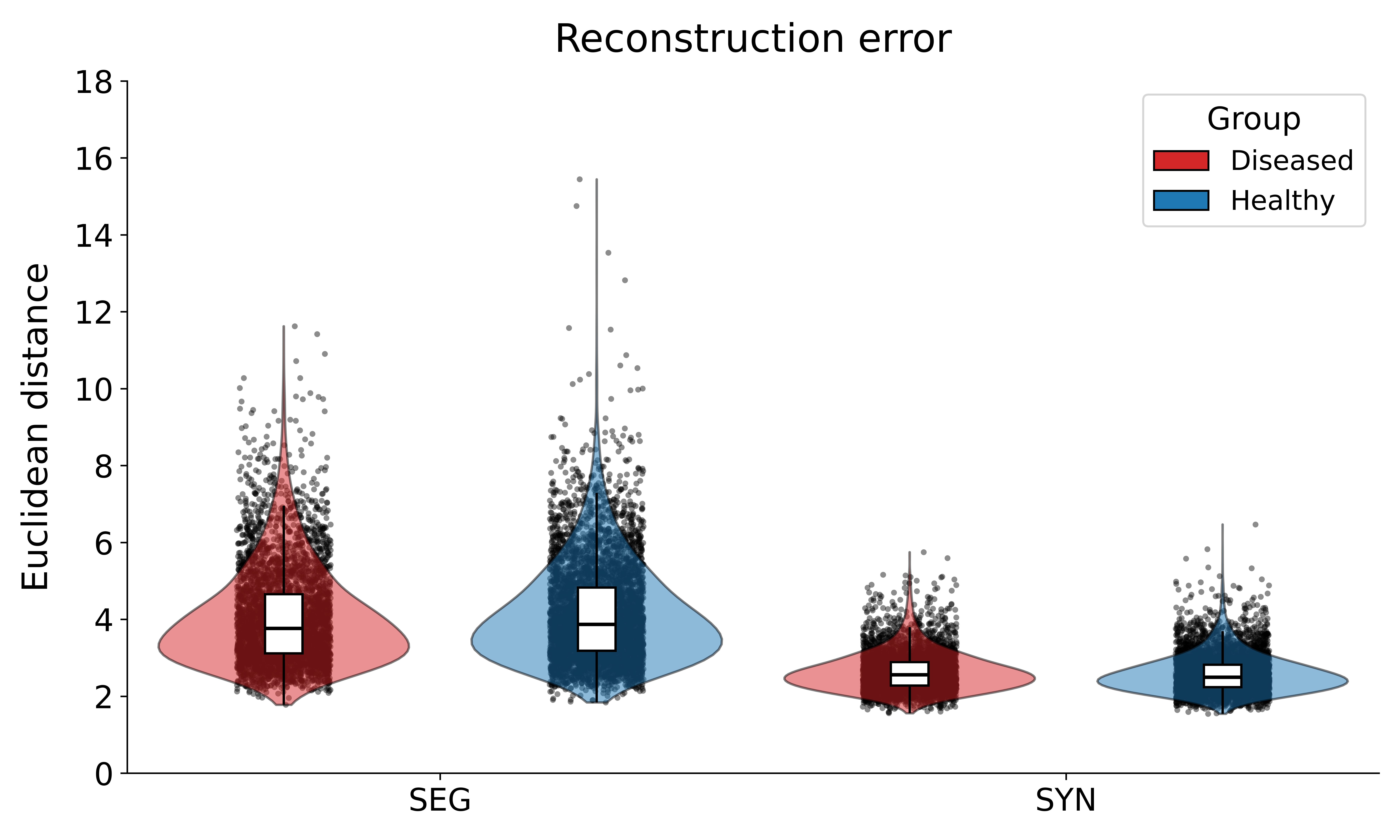}
    \caption{Comparison of ED values between REF meshes and SYN/SEG predictions in healthy and diseased samples. Violin plots show the distribution of values, while boxplots indicate medians and interquartile ranges.}
    \label{fig:res2}
\end{figure}

The numerical results for the second experiment, including contour-based synthetic (SYN) and real (SEG) predictions, are presented in Table~\ref{tab:res2}. Since all metrics are computed relative to the reference mesh, the highest performance in both classification and regression tasks was observed for the SYN group. This result was expected, as the input slices used for inference optimization in the SYN group are perfectly consistent with the underlying reference meshes. Notably, approximately 62--65\% of the Euclidean distance error was attributable to the domain gap between the reference meshes used for training and the misaligned contours used during testing. Fig.~\ref{fig:res2} further illustrates this difference by showing the distribution of the ED values. A clear domain gap between the SYN and SEG groups is evident, primarily due to differences in how the input slices were constructed and their degree of similarity to the training data. This discrepancy negatively impacts the model's performance on real cases, whereas the SYN group benefits from a closer alignment with the training distribution.

\begin{table}[H]
    \centering
    \caption[Exp1-table]{Results of the regression task for the first experiment. The reconstruction error is shown using Euclidean distance (ED), root mean square error (RMSE), symmetric and aysymmetric Chamfer distance (CD) comparing SEG and REF meshes to cvi42 contours, for both cardiovascular diseased (CVD) and no cardiovascular diseased (No-CVD) cases. All results are shown as avg$\pm$std (mm).}
    \resizebox{0.9\textwidth}{!}{
    \renewcommand{\arraystretch}{1.3}
    \begin{tabular}{|ccccc|}
    \hline
     & \multicolumn{4}{c|}{Regression Metrics} \\
    \hline
     & ED & RMSE & CD(sym) & CD(asym)\\
    \hline
     \multicolumn{5}{|c|}{CVD Group Set (n= 4549)} \\
    \hline
        SEG & 2.51 ± 0.33 & 2.96 ± 0.44 & 3.16 ± 0.30 & 2.51 ± 0.33 \\
        REF & 2.31 ± 0.36 & 2.81 ± 0.50 & 2.98 ± 0.34 & 2.31 ± 0.36 \\
    \hline
    \multicolumn{5}{|c|}{No-CVD Group Set (n= 5576)} \\
    \hline
        SEG  & 2.53 ± 0.33 & 2.99 ± 0.46 & 3.15 ± 0.32 & 2.53 ± 0.33\\
        REF  & 2.24 ± 0.38 & 2.73 ± 0.53 & 2.91 ± 0.35 & 2.24 ± 0.38\\
    \hline
    \end{tabular}}
    \label{tab:res1}
\end{table}

\begin{table}[H]
    \centering
    \caption[Exp2-table]{Results of the classification and regression task for the second experiment. The classification error is expressed using the Dice metric (F1-score) for the left ventricular myocardium (LVM) and right ventricular myocardium (RVM) labels compared against labels from the reference mesh. The reconstruction error is shown using Euclidean distance (ED), root mean square error (RMSE), symmetric and aysymmetric Chamfer distance (CD), comparing SYN/SEG predictions to REF meshes. All results are shown as avg$\pm$std (mm).}
    \resizebox{\textwidth}{!}{
    \renewcommand{\arraystretch}{1.3}
    \begin{tabular}{|ccccccc|}
    \hline
     & \multicolumn{2}{c}{Classification Metrics} & \multicolumn{4}{c|}{Regression Metrics} \\
    \hline
     & DICE LVM & DICE RVM & ED & RMSE & CD(sym) & CD(asym)\\
    \hline
     \multicolumn{7}{|c|}{CVD Group Set (n= 4549)} \\
    \hline
        SEG & 0.91 ± 0.04 & 0.87 ± 0.04 & 4.06 ± 1.29 & 4.47 ± 1.37 & 2.10 ± 0.30 & 2.09 ± 0.32\\
        SYN & 0.95 ± 0.02 & 0.91 ± 0.03 & 2.63 ± 0.50 & 2.96 ± 0.56 & 1.76 ± 0.18 & 1.75 ± 0.18\\
    \hline
    \multicolumn{7}{|c|}{No-CVD Group Set (n= 5576)} \\
    \hline
        SEG & 0.91 ± 0.04 & 0.87 ± 0.05 & 4.14 ± 1.30 & 4.55 ± 1.40 & 2.13 ± 0.32 & 2.12 ± 0.33\\
        SYN & 0.96 ± 0.02 & 0.91 ± 0.03 & 2.58 ± 0.49 & 2.89 ± 0.55 & 1.74 ± 0.20 & 1.74 ± 0.21\\
    \hline
    \end{tabular}}
    \label{tab:res2}
\end{table}
\vspace{0.8cm}

Table \ref{tab:res3} presents the volume estimations derived from our network’s predictions for both the SEG and SYN groups, compared against the volumes estimated from the REF meshes for both CVD and non-CVD groups. Additional results and further analysis can be found in the Appendix.

\begin{table}[h]
    \centering
    \caption{Volume estimated from the network (SEG/SYN groups) and from standard methods (REF). CVD: Cardiovascular diseased cases; LV: Left ventricle; RV: Right ventricle.}
    \resizebox{\textwidth}{!}{
    \renewcommand{\arraystretch}{1.3}
    \begin{tabular}{|c|c|c|c|c|c|}
        \hline
         & & LV VOL (mL) & LV MASS (g) & RV VOL (mL) & RV MASS (g) \\
        \hline
        \multirow{3}{*}{SEG} 
        & CVD     & 147.56 ± 32.44 & 122.44 ± 23.93 & 155.22 ± 36.72 & 52.05 ± 10.79  \\
        & No-CVD  & 149.54 ± 45.79 & 117.92 ± 32.53 & 160.49 ± 56.92 & 51.98 ± 16.56  \\
        \hline
        \multirow{3}{*}{SYN} 
        & CVD     & 144.35 ± 30.57 & 136.04 ± 25.35 & 151.00 ± 31.77 & 53.13 ± 10.17 \\
        & No-CVD  & 144.50 ± 45.08 & 132.47 ± 38.27 & 152.14 ± 49.91 & 52.66 ±16.01  \\
        \hline
        \multirow{3}{*}{REF}      
        & CVD     & 155.55 ± 35.47 & 118.02 ± 25.15 & 153.35 ± 34.60 & 50.36 ± 10.47  \\
        & No-CVD  & 155.90 ± 50.56 & 113.80 ± 35.64 & 157.49 ± 54.17 & 51.17 ± 16.28 \\
        \hline
    \end{tabular}}
    \label{tab:res3}
\end{table}

In the following figures, we display and quantify the meshes obtained in the SEG group. For the SYN predictions, please refer to the Supplementary Material.\\

\begin{figure}[H]
    \centering
    \includegraphics[width=0.85\textwidth]{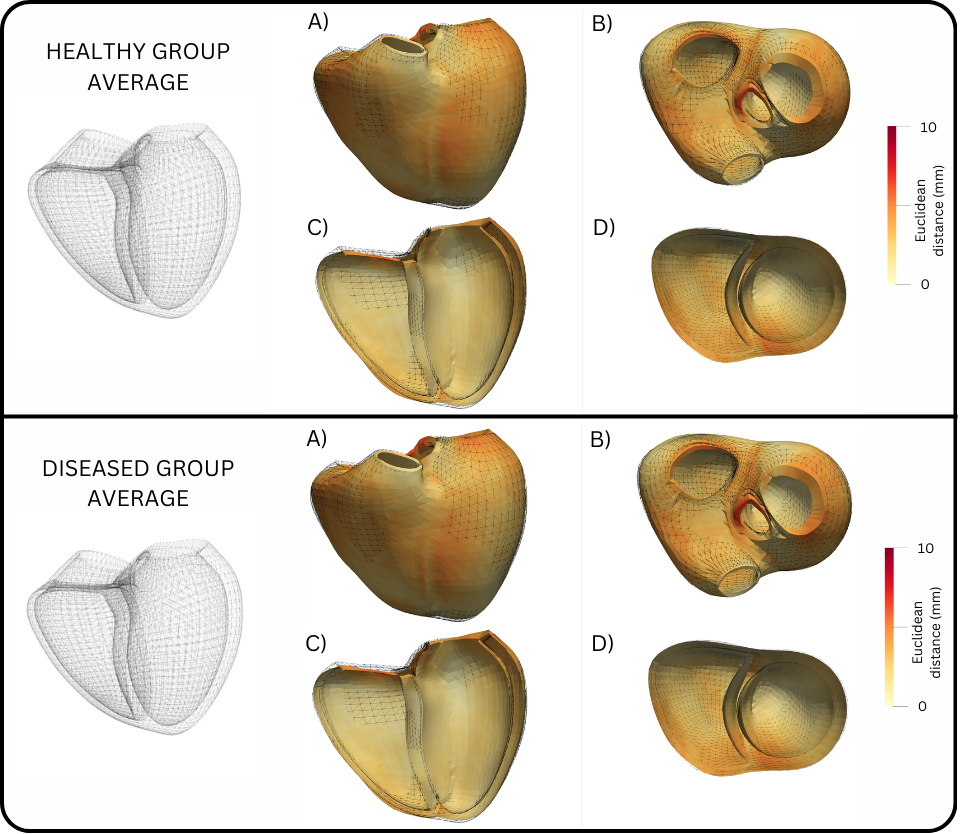}
    \caption[SEG average mesh]{Results of the regression task for the SEG mesh predictions. The average shape is obtained for both the No-CVD and CVD group sets by averaging the predicted points and the reference points. On the left, the reference mesh is shown. On the right, the predicted mesh is shown with a coloured colormap based on Euclidean distance from reference mesh, whereas the reference mesh is overlayed with a wireframe. All shapes are equally scaled to preserve size information.}
    \label{fig:res3}
\end{figure}

Fig.~\ref{fig:res3} illustrates the prediction trends of the model, highlighting the average results obtained across both the CVD and No-CVD test sets. On the left panel, we display the average reference mesh in wireframe (representing the population mean of vertex locations rather than individual cases), and on the right panel, the corresponding average prediction. The heatmap overlays the average vertex-wise error relative to the reference mesh, indicating regions with higher or lower prediction errors. Overall, the error distribution is smooth, with slightly elevated errors observed near the valve plane — the most challenging anatomical region with UVC similarities — and on the epicardial surface. The wireframe provides a visual reference of the averaged anatomical structure across the test population.

\begin{figure}[H]
    \centering
    \includegraphics[height=0.85\textheight]{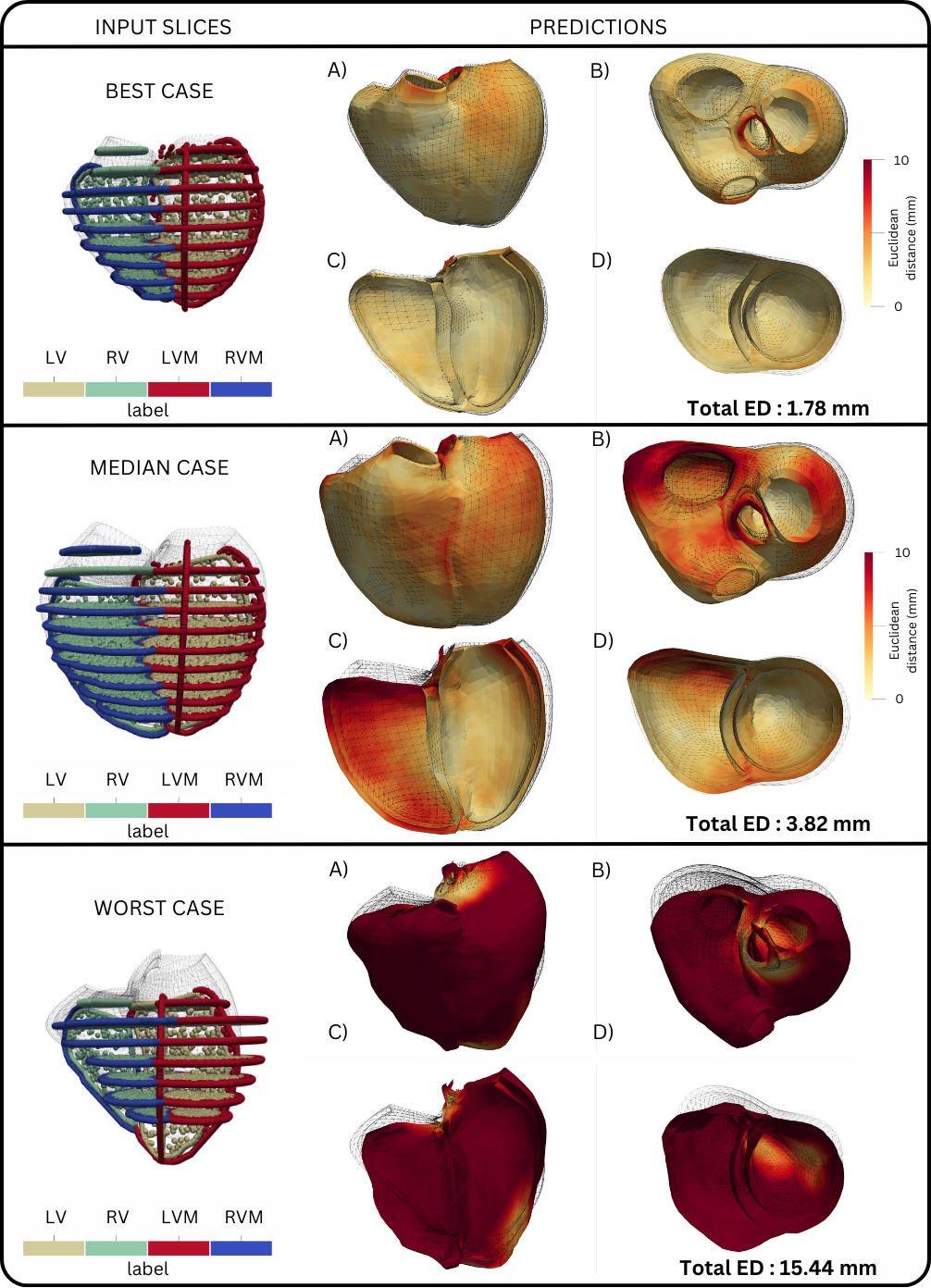}
    \caption[SEG 3 meshes]{Results of the regression task for the SEG meshes. Meshes are shown for best, median and worst cases overall (calculated over both No-CVD and CVD groups). Best case is from the CVD group, median and worst are from the No-CVD group. On the right, the predicted mesh is shown with a coloured colormap based on Euclidean distance from reference mesh, whereas the reference mesh is overlayed with a wireframe. All shapes are equally scaled to preserve size information.}
    \label{fig:res4}
\end{figure}

Fig.~\ref{fig:res4} summarizes the quality of the predicted meshes. The best case shows near-perfect alignment with the reference mesh, achieving an average Euclidean distance (ED) of 1.78~mm. In this case, most errors are localized around the valve plane. The median case exhibits moderate errors, particularly in the right ventricle and around the tricuspid valve plane. This may be attributed to the absence of contour information for the most basal slice at the level of the tricuspid valve. In the worst case, significant misalignment of the input slices results in a severely deformed predicted mesh.

Finally, we conducted an ablation study to assess the influence of each slice on the predicted meshes. The numerical results for all cases are presented in Table~\ref{tab:res4}. These results show that removing the LAX 3CH slice led to slightly better predictions, suggesting that this slice may introduce more inconsistencies with other slices in the contours.

In the SYN group (Appendix~B, Fig.~\ref{fig:app-res3}), both the segmentation and regression tasks achieved the best performance when the full set of slices was used. This outcome is expected, as the synthetic slices are perfectly consistent with the reference meshes.

\begin{figure}[H]
    \centering
    \includegraphics[width=\textwidth]{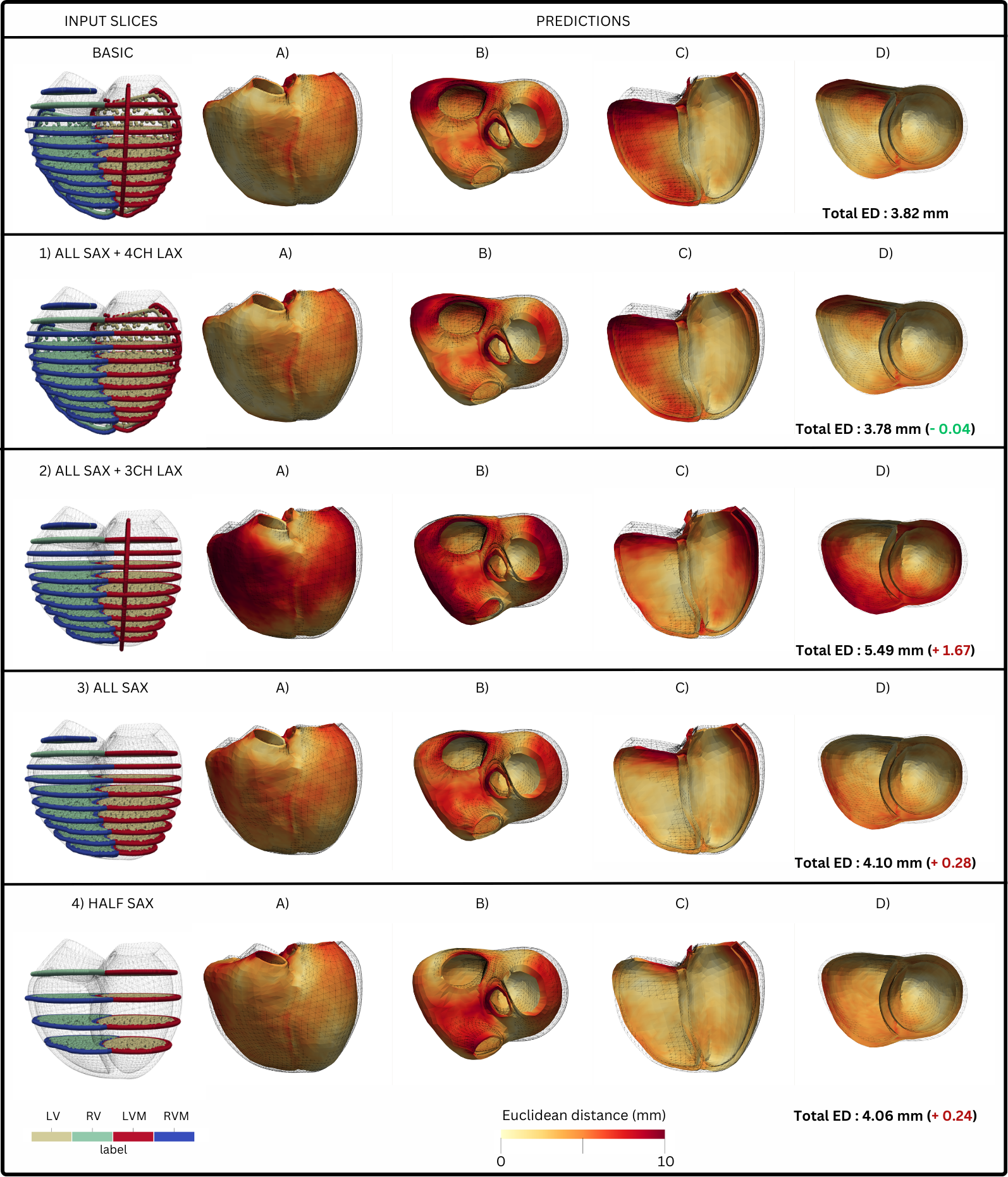}
    \caption[SEG ablation median]{Predictions of ablation study for the median case in the real contour group. On the left, the input points used at inference optimisation in all the ablation cases are displayed. On the right, the predicted mesh is shown with a coloured colormap based on Euclidean distance from reference mesh, whereas the reference mesh is overlayed with a wireframe. The Euclidean distance from the reference mesh is shown with its relative difference to the full experiment (result in Fig.\ref{fig:res4}- median case). All shapes are equally scaled to preserve size information.}
    \label{fig:res5}
\end{figure}

Fig.~\ref{fig:res5} illustrates the predicted meshes obtained from the ablation study. Notably, the overall prediction error increases substantially in case (2), where the LAX 4CH slice was removed.

\begin{table}[H]
    \centering
    \caption{Ablation study results for real contour-based meshes. CVD: Cardiovascular diseased cases; LAX: Long-axis; SAX: short-axis; LVM: left ventricular mass; RVM: right ventricular mass; ED: Euclidean distance; RMSE: root mean square error.}
    \resizebox{\textwidth}{!}{
    \renewcommand{\arraystretch}{1.3}
    \begin{tabular}{cccc|cc|cc}
        \hline
        \multicolumn{8}{c}{\textbf{CVD Group Set (n= 4549)}} \\
        \hline
        \multicolumn{4}{c|}{\textbf{Segmentation Views}} & \multicolumn{2}{c|}{\textbf{ Classification Metrics}} & \multicolumn{2}{c}{\textbf{Regression Metrics}} \\
        \textbf{LAX 3 CH} & \textbf{LAX 4 CH} & \textbf{HALF SAX} & \textbf{ALL SAX} & \textbf{DICE LVM} & \textbf{DICE RVM} & \textbf{ED} & \textbf{RMSE} \\
        \hline
        \cmark & \cmark & \xmark & \cmark & \textbf{0.91 ± 0.04} & \textbf{0.87 ± 0.04} & 4.06 ± 1.29 & 2.34 ± 0.74 \\
        \xmark & \cmark & \xmark & \cmark & 0.90 ± 0.04 & 0.87 ± 0.04 & \textbf{3.88 ± 1.20} & \textbf{2.24 ± 0.69}\\
        \cmark & \xmark & \xmark & \cmark & 0.90 ± 0.04 & 0.86 ± 0.05 & 4.16 ± 1.29 & 2.40 ± 0.75 \\
        \xmark & \xmark & \xmark & \cmark & 0.88 ± 0.05 & 0.86 ± 0.05 & 3.94 ± 1.14 & 2.27 ± 0.66 \\
        \xmark & \xmark & \cmark & \xmark & 0.86 ± 0.05 & 0.78 ± 0.07 & 4.33 ± 1.13 & 2.50 ± 0.65 \\
        \hline
        \multicolumn{8}{c}{\textbf{No-CVD Group Set (n= 5576)}} \\
        \hline
        \multicolumn{4}{c|}{\textbf{Segmentation Views}} & \multicolumn{2}{c|}{\textbf{ Classification Metrics}} & \multicolumn{2}{c}{\textbf{Regression Metrics}}  \\
        \textbf{LAX 3 CH} & \textbf{LAX 4 CH} & \textbf{HALF SAX} & \textbf{ALL SAX} & \textbf{DICE LVM} & \textbf{DICE RVM} & \textbf{ED} & \textbf{RMSE} \\
        \hline
        \cmark & \cmark & \xmark & \cmark & \textbf{0.91 ± 0.04} & \textbf{0.87 ± 0.05} & 4.14 ± 1.30 & 2.39 ± 0.75 \\
        \xmark & \cmark & \xmark & \cmark & 0.89 ± 0.04 & 0.87 ± 0.05 & \textbf{4.00 ± 1.23} & \textbf{2.31 ± 0.71} \\
        \cmark & \xmark & \xmark & \cmark & 0.90 ± 0.04 & 0.86 ± 0.05 & 4.23 ± 1.27 & 2.44 ± 0.73 \\
        \xmark & \xmark & \xmark & \cmark & 0.88 ± 0.05 & 0.86 ± 0.05 & 4.06 ± 1.20 & 2.35 ± 0.69 \\
        \xmark & \xmark & \cmark & \xmark & 0.85 ± 0.05 & 0.77 ± 0.08 & 4.46 ± 1.23 & 2.57 ± 0.71 \\
        \hline
    \end{tabular}}
    \label{tab:res4}
\end{table}

\section{Discussion}

In this paper, we introduce a novel application of Neural Implicit Functions (NIFs) for reconstructing 3D biventricular meshes from labeled slices. To achieve this, we leverage UVC maps, computed from the novel CobivecoX method \cite{pankewitz2024universal} which provide a consistent reference system for heart locations and allow us to predict whole biventricular meshes including valve planes and the outflow tracts. In addition, while existing methods for computing UVCs require a high-resolution 3D input mesh to solve partial differential equations, our approach can compute NIHC maps directly from sparse segmentation data. Given the NIHCs, a mesh can be reconstructed at any resolution by assigning NIHC values to each vertex.

Our model is trained on a large dataset of 5,000 cardiac meshes, previously obtained in \cite{burns2024genetic}. We evaluated the method across two groups, CVD and No-CVD, using both synthetic slices (free from spatial inconsistencies) and real contours, which exhibit spatial inconsistencies due to segmentation variability and misalignment.  We consider the discrepancy between synthetic and real contours to be a primary factor influencing the performance of our model, and thus a domain gap between training and testing data. In contrast, we observe that anatomical variability between the CVD and No-CVD groups has a comparatively smaller impact on model performance.

In our first experiment (Table~\ref{tab:res1} and Fig.~\ref{fig:res1}), we show that the error between the SEG predictions and the original cvi42 contours is comparable to the error between the REF meshes and the contours. Furthermore, the standard deviation of the errors across the population is generally smaller for the SEG predictions. This experiment supports the validity of the proposed method and suggests its potential to outperform standard approaches in scenarios where the alignment problem is resolved.

Our method achieves a Dice score above 0.9 for the left ventricular myocardium (LVM) in both the SEG and synthetic (SYN) groups, and between 0.87 and 0.91 for the right ventricular myocardium (RVM) (Table~\ref{tab:res2}). As Dice scores are computed at the mesh vertices—specifically on the epicardial and endocardial surfaces—these results demonstrate high reconstruction accuracy. The performance differences that can be observed in the regression metrics and in Fig.~\ref{fig:res2} are likely attributable to the aforementioned domain gap between real and synthetic data. In particular, while training was performed using points sampled within a 3D bounding box around the reference meshes, the SEG contours used at test time can exhibit noise and inter-slice misalignment. In contrast, the SYN group slices are directly sampled from the reference meshes and therefore exhibit perfect spatial consistency. To further illustrate this effect, we provide a supplementary figure (Fig.~\ref{fig:app-res8}) showing the five worst-performing cases in both the SYN and SEG groups. These outliers are primarily attributable to strong misalignment in the input data, rather than limitations of the proposed method itself. Notably, the worst cases in the SEG group exhibit substantially higher error metrics than those in the SYN group, a trend that is also reflected in the violin plots (Fig.~\ref{fig:res2}).

The average shapes shown in Fig.~\ref{fig:res3} (and Appendix~B, Fig.~\ref{fig:app-res1}) demonstrate that our method produces spatially consistent meshes without significant localized error accumulation. Here, we define significant localized error as spatially contiguous regions exhibiting reconstruction errors substantially above the typical contour-to-mesh fitting discrepancy (median $\approx$ 2.5 mm, ref. Fig.\ref{fig:res2}). The majority of the cardiac surface exhibits errors below 3 mm, with values generally ranging between 0 and 10 mm and no large contiguous regions of higher error. Over 60\% of vertices exhibit errors below 3 mm, and fewer than 2\% exceed 5 mm. 
As expected, higher errors are primarily confined to the valve plane, which is known to be challenging to reconstruct due to the absence of labeled slices in standard MR acquisitions and increased misalignment, resulting in limited or incorrect partial supervision during inference optimization. Moreover, this region corresponds to an anatomical area where the definition and uniqueness of UVCs are inherently more ambiguous, making it more susceptible to reconstruction inaccuracies \cite{pankewitz2024universal}.

NIF–based approaches are inherently flexible in that, unlike CNN-based methods, they do not require a fixed set of images or predefined spatial grids at inference time. This property is well established in prior work, including Amiranashvili et al. and our own previous studies \cite{amiranashvili2022learning,amiranashvili2024learning,muffoletto2023neural} and it derives from training the model on continuous spatial coordinates rather than discrete frames. In this work, we leverage this characteristic in the specific context of cardiac mesh reconstruction, where the NIF is queried at the spatial locations corresponding to a template mesh. Although in our experiments the primary application is the reconstruction of 3D cardiac meshes, and thus we restrict NIF queries to predefined UVC locations corresponding to the template mesh, the pipeline naturally enables future extensions in which the NIF is queried at arbitrary spatial locations, for example in digital twin or patient-specific simulation settings.

This inherent flexibility of NIFs is reflected in our ablation study, where the method maintains robust performance even under a substantial reduction in input data. In particular, experiment 4 (“HALF SAX”) shows that accurate reconstructions can still be obtained with only half of the short-axis slices. While this result is encouraging, it likely corresponds to a median scenario with well-aligned slices and anatomy closely matching the learned shape prior. Consequently, such performance may not generalize to subjects that strongly deviate from the prior distribution, especially in pathological cases with pronounced local remodeling, where additional long-axis (LAX) views become critical to capture fine structural variations.

The ablation study further shows that the most accurate reconstructions are obtained when combining all SAX slices with the 4CH LAX view (Table \ref{tab:res4}). In contrast, configurations relying on the 3CH view appear less effective, likely due to its higher susceptibility to misalignment, which can negatively impact reconstruction quality. This effect is illustrated in Fig.~\ref{fig:res5}, where misaligned slices in the SEG test group lead to a degradation in the overall Euclidean distance (ED). In particular, when the 3CH LAX slice is included while the 4CH slice is removed (case 2), the prediction error increases substantially (+1.67~mm). This behavior is observed for real contour–based predictions (SYN group), but not for the SEG group (Table \ref{tab:app-table1}, Fig.\ref{fig:app-res3}), where incorporating additional slices consistently improves performance. In the SEG setting, all slices are synthetically aligned and therefore provide complementary supervision, resulting in more intuitive improvements as the number of points used during inference optimization increases. These findings further highlight the domain gap between synthetic and real data, with errors in real contour–based reconstructions being primarily attributable to slice misalignment.

Finally, a strength of our method was the robustness shown when applied to a CVD test set, despite being trained exclusively on No-CVD subjects. The differences in predictions between the CVD and No-CVD groups are generally negligible. Interestingly, more outliers are observed in the No-CVD test set, likely due to the presence of cases with very large volumes (Table \ref{tab:res2} shows higher average values), which can sometimes pose a challenge for the network.

To mitigate the effects of slice misalignment and to encourage anatomically plausible reconstructions, we introduced a shape regularization term (Section~2.3.2). In a separate experiment on the SYN contour group, improved performance was obtained by setting $\lambda_{\text{reg}} = 10^{-4}$, consistent with the original DeepSDF formulation~\cite{park2019deepsdf}. However, due to the comparatively low inter-subject shape variability of cardiac anatomy (relative to generic object classes in computer vision) and the presence of noise during inference-time optimization, we observed that stronger regularization was more effective. This prevents predictions from deviating significantly from plausible heart shapes and reduces overfitting to inconsistent input slices. 

Looking ahead, the impact of slice misalignment could be further reduced by applying a pre-alignment step to real contour data, for instance using the approach proposed by Sinclair et al.~\cite{sinclair2017fully}. In addition, motivated by our ablation study showing a clear degradation in performance when misaligned slices are included, the proposed framework could be complemented by an automated quality-control mechanism to detect and downweight (or discard) low-quality or inconsistent input slices.

Beyond these measures, the flexibility of the inference step opens the possibility of adopting region- or constraint-aware formulations to enable more localized improvements based on the available input data. For example, one could first perform inference using only SAX slices to query the full ventricular domain, and subsequently incorporate selected LAX slices while restricting the prediction to specific anatomical regions (e.g., left-ventricular points), which can be reliably identified thanks to the intrinsic properties of the UVC system. This could be further combined with improved neural implicit strategies that model structures interdependently \cite{le2025pairwise, talabot2025partsdf}.

Beyond misalignment, other limitations of the proposed approach are largely attributable to the nature of the input data. First, the right ventricular (RV) epicardial surface was not directly available in the input segmentations and was therefore estimated by assuming a constant RV myocardial thickness of 3mm. This value corresponds to the reported mean normal RV wall thickness in adults~\cite{foale1986echocardiographic} and was required for the computation of the UVCs. While this assumption is reasonable for healthy cohorts, it may not hold in pathological cases such as RV hypertrophy. Future work should therefore investigate the ability of the proposed framework to reconstruct variable RV wall thicknesses and to relax this prior when sufficient information is available. Second, the use of mesh-derived contours introduces an inherent level of approximation, as these contours are not directly observed in the images. Nevertheless, they are typically more consistent and topologically coherent than contours extracted independently from 2D images, particularly across slices and subjects. This trade-off promotes smoother and anatomically plausible meshes during training, at the cost of deviating from image-derived boundaries. More fundamentally, the lack of true 3D ground-truth anatomy when fitting volumetric meshes to sparse 2D CMR acquisitions remains a well-recognized limitation in cardiac shape modeling and segmentation. Prior work has shown that segmentation and fitting accuracy are inherently constrained by this ambiguity~\cite{mauger2019right}, and addressing this limitation will likely require richer imaging protocols or the integration of stronger anatomical priors in future studies.

\section{Conclusion}

In this study, we have developed a novel NIF-based method capable of accurately reconstructing 3D biventricular surface anatomies from sparse and misaligned cine MRI contours, leveraging Neural Ventricular Coordinates. Our framework consists of a weakly supervised classification task, based on anatomical labels, and a regression task, which can accurately predict meshes at any resolution. Our approach significantly reduces the time required for full mesh reconstruction, from approximately 60 seconds with traditional mesh fitting methods to 5-15 seconds (combined inference-time optimization and prediction).  
The method demonstrates strong robustness to a varying number of input slices at inference time, as well as to significant noise and spatial misalignments in the segmentation contours. The main limitation of the current approach is a high degree of sensitivity to slice misalignment, which can be addressed by adding a pre-alignment step to the input data, or by explicitly training the network with noise to enhance robustness. Future work will aim to enhance the network architecture, extend the approach to additional cardiac substructures and the full cardiac cycle, and further assess its performance across a broader range of cardiac pathologies.

\section{Funding}
This research was conducted using the UK Biobank Resource under Application Number 2964. The work uses data provided by patients and collected by the NHS as part of their care and support. This research used data assets made available by National Safe Haven as part of the Data and Connectivity National Core Study, led by Health Data Research UK in partnership with the Office for National Statistics and funded by UK Research and Innovation (grant ref MC\_PC\_20029). Barts Charity (G- 002346) contributed to fees required to access UK Biobank data [access application \#2964]. MM and AY acknowledge core funding from the Wellcome/EPSRC Centre for Medical Engineering [WT203148/Z/16/Z] and The London Medical Imaging \& AI Centre for Value Based Healthcare. MM was supported by EPSRC Centre for Doctoral Training in Smart Medical Imaging (EP/S022104/1) and by Siemens Healthineers. AY CM and AM are supported by National Institutes of Health R01HL121754. SEP acknowledges the British Heart Foundation for funding the manual analysis to create a cardiovascular magnetic resonance imaging reference standard for the UK Biobank imaging resource in 5000 CMR scans (www.bhf.org.uk; PG/14/89/31194), as well as from the “SmartHeart” EPSRC programme grant (www.nihr.ac. uk; EP/P001009/1). SEP has received funding from the European Union’s Horizon 2020 research and innovation programme under grant agreement No 825903 (euCanSHare project). SEP acknowledge the support of the National Institute for Health and Care Research Barts Biomedical Research Centre (NIHR203330); a delivery partnership of Barts Health NHS Trust, Queen Mary University of London, St George’s University Hospitals NHS Foundation Trust and St George’s University of London.

\section*{Declaration of competing interests}
ADM is required to disclose that he is co-founder and advisor to Insilicomed, Inc. and Vektor Medical, Inc. Neither company was involved in this study in any way. The other authors declare that they have no known competing financial interests or personal relationships that could have appeared to influence the work reported in this paper.

\bibliographystyle{elsarticle-num} 
\bibliography{bib_list.bib}

\newpage
\appendix

\section{Implementation Details}
\label{app:implementation}

Each mesh was aligned by transforming its coordinates to the same Cartesian "cardiac" coordinate space, using an affine transformation matrix in homogeneous coordinates $(x, y, z, 1)$, denoted as $A = [R | -O]$. Here, the rotation matrix $R$ was decomposed into components [XA, YA, ZA] defined as follows:

\begin{itemize}
    \item The LV apex, centroids of the mitral and tricuspid valves were identified.
    \item The long axis (LAX) was defined as the line from the mitral valve centroid (MVC) to the LV apex (LVA).
    \item The origin of the cardiac coordinates ($O$) was set at the midpoint of the line segment from MVC to LVA ($O = (MVC + LVA) / 2$).
    \item The Z axis (ZA) was defined as the normalised vector pointing from O to LVA ($ZA = LVA - O$).
    \item The Y axis (YA) was defined as the normalised vector perpendicular to ZA, oriented left to right on the four-chamber (4CH) plane. Initially, YA was initialised as the vector from O to the tricuspid valve centroid (TVA), then adjusted as:
    \begin{itemize}
        \item $YAP = \text{dot}(YA, ZA) \cdot ZA$ (projection onto ZA),
        \item $YA = YA - YAP$ (resulting in YA perpendicular to ZA).
    \end{itemize}
    \item The X axis (XA) was determined as the cross product of YA and ZA.
\end{itemize}

\begin{figure}[H]
    \centering
    \includegraphics[width=\textwidth]{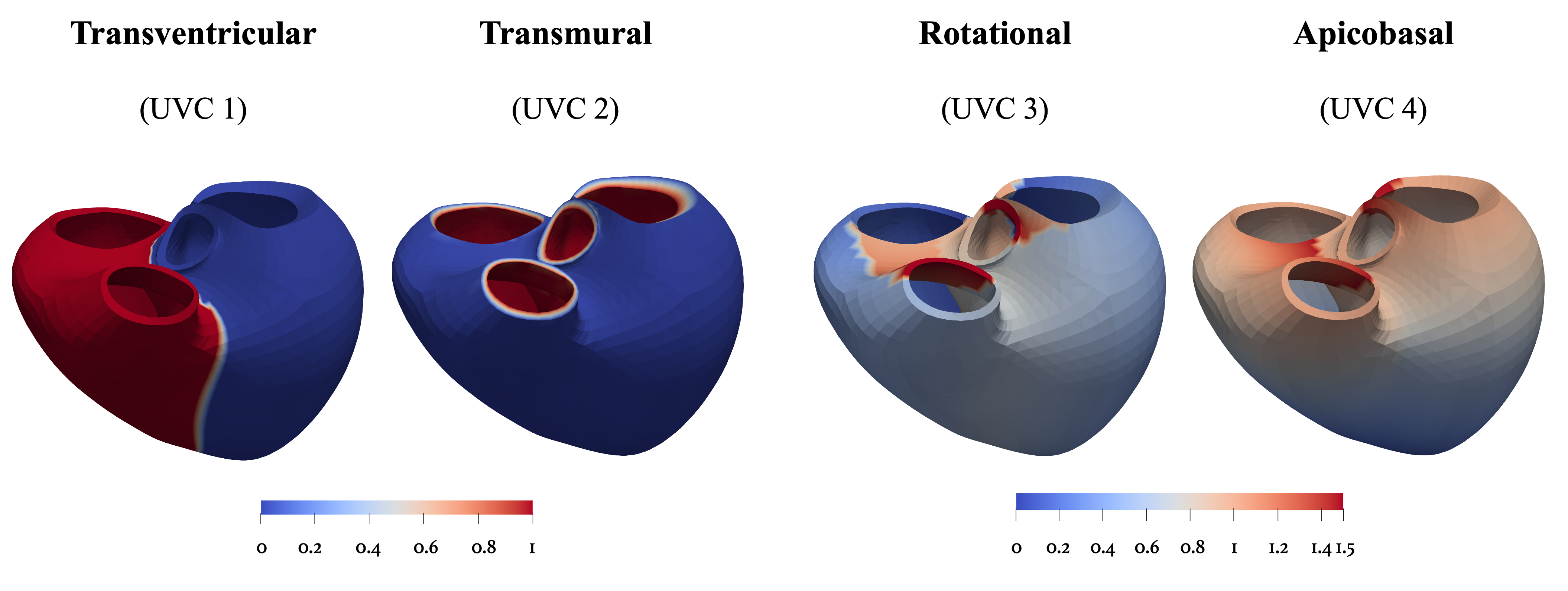}
    \caption[UVC-CobivecoX]{CobivecoX UVC maps calculated for the mean Biobank mesh, using the method described in  \cite{pankewitz2024universal}.}
    \label{fig:app-uvc}
\end{figure}

\section{Extended Results}
\label{app:results}

\begin{figure}[H]
    \centering
    \includegraphics[width=0.85\textwidth]{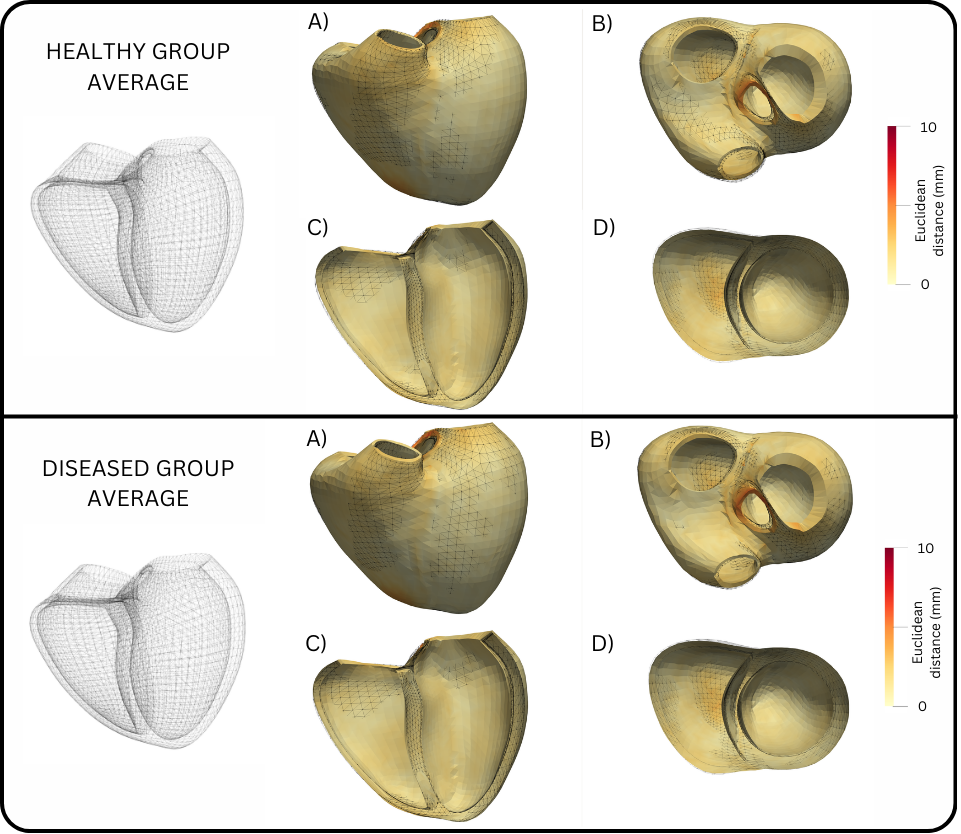}
    \caption[SYN average mesh]{Results of the regression task for the synthetic (SYN) contours group. The average shape is obtained for both the No-CVD and CVD group sets by averaging the predicted points and the reference points. On the left, the reference mesh is shown. On the right, the predicted mesh is shown with a coloured colormap based on Euclidean distance from reference mesh, whereas the reference mesh is overlayed with a wireframe. All shapes are equally scaled to preserve size information.}
    \label{fig:app-res1}
\end{figure}

\begin{figure}[H]
    \centering
    \includegraphics[height=0.85\textheight]{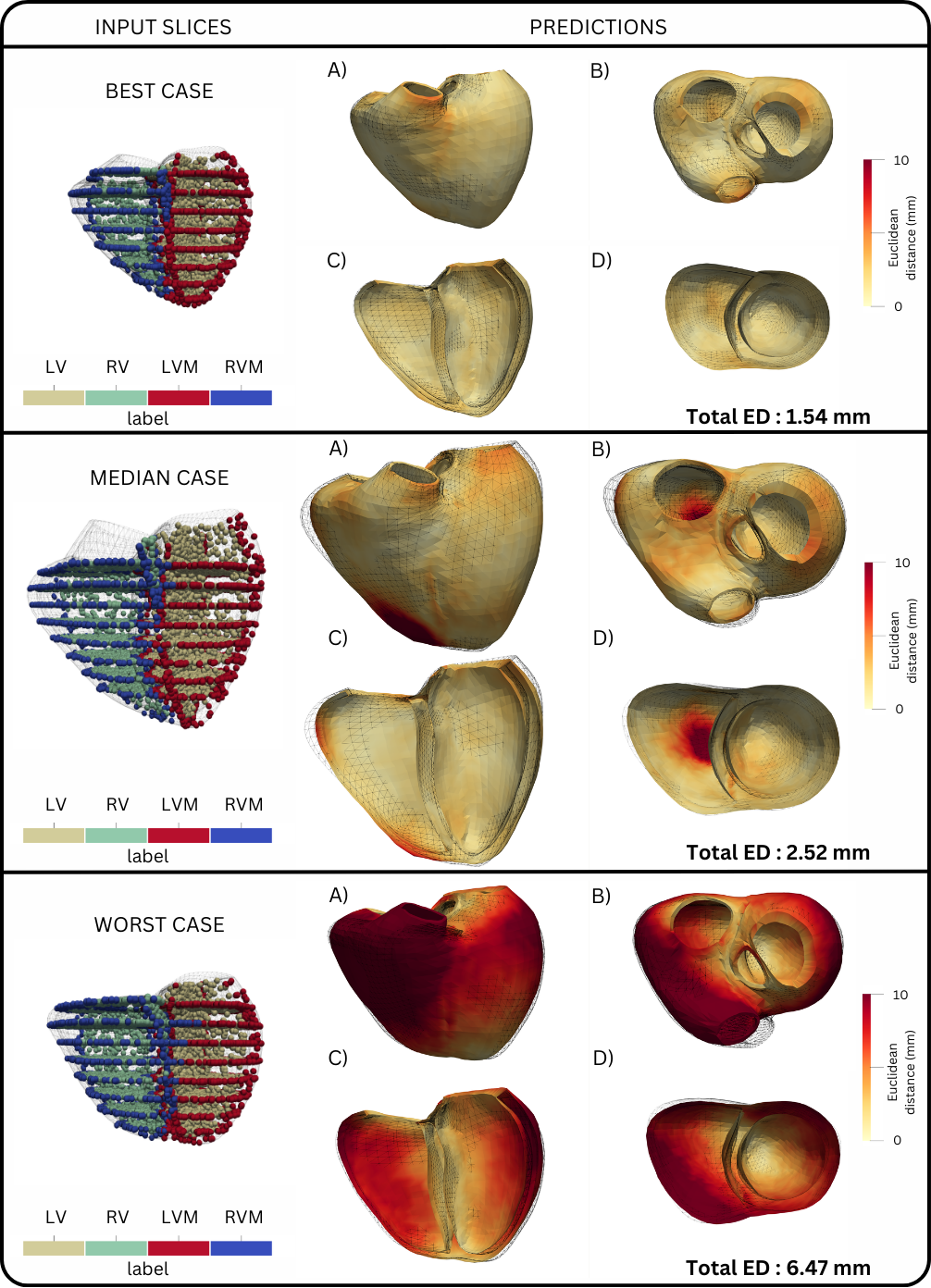}
    \caption[SYN 3 meshes]{Results of the regression task for the synthetic (SYN) contours group. Meshes are shown for best, median and worst cases overall (calculated over both No-CVD and CVD groups). All cases are from the No-CVD group. On the left, the input points used at inference optimisation are overlayed on the reference mesh. On the right, the predicted mesh is shown with a coloured colormap based on Euclidean distance from reference mesh, whereas the reference mesh is overlayed with a wireframe. All shapes are equally scaled to preserve size information.}
    \label{fig:app-res2}
\end{figure}

\begin{table}[H]
    \centering
    \caption{Ablation study results for synthetic contour-based meshes.}
    \resizebox{\textwidth}{!}{
    \renewcommand{\arraystretch}{1.3}
    \begin{tabular}{cccc|cc|cc}
        \hline
        \multicolumn{8}{c}{\textbf{CVD Group Set (n= 4549)}} \\
        \hline
        \multicolumn{4}{c|}{\textbf{Segmentation Views}} & \multicolumn{2}{c|}{\textbf{ Classification Metrics}} & \multicolumn{2}{c}{\textbf{Regression Metrics}} \\
        \textbf{LAX 3 CH} & \textbf{LAX 4 CH} & \textbf{HALF SAX} & \textbf{ALL SAX} & \textbf{DICE LVM} & \textbf{DICE RVM} & \textbf{ED} & \textbf{RMSE} \\
        \hline
        \cmark & \cmark & \xmark & \cmark & \textbf{0.95 ± 0.02} & \textbf{0.91 ± 0.03} & \textbf{2.63 ± 0.50} & \textbf{1.52 ± 0.29} \\
        \xmark & \cmark & \xmark & \cmark & 0.95 ± 0.03 & 0.91 ± 0.03 & 2.64 ± 0.51 & 1.53 ± 0.29 \\
        \cmark & \xmark & \xmark & \cmark & 0.95 ± 0.03 & 0.91 ± 0.03 & 2.67 ± 0.52 & 1.54 ± 0.30 \\
        \xmark & \xmark & \xmark & \cmark & 0.94 ± 0.03 & 0.91 ± 0.03 & 2.69 ± 0.54 & 1.55 ± 0.31 \\
        \xmark & \xmark & \cmark & \xmark & 0.93 ± 0.04 & 0.88 ± 0.04 & 3.03 ± 0.66 & 1.75 ± 0.38 \\
        \hline
        \multicolumn{8}{c}{\textbf{No-CVD Group Set (n= 5576)}} \\
        \hline
        \multicolumn{4}{c|}{\textbf{Segmentation Views}} & \multicolumn{2}{c|}{\textbf{ Classification Metrics}} & \multicolumn{2}{c}{\textbf{Regression Metrics}}  \\
        \textbf{LAX 3 CH} & \textbf{LAX 4 CH} & \textbf{HALF SAX} & \textbf{ALL SAX} & \textbf{DICE LVM} & \textbf{DICE RVM} & \textbf{ED} & \textbf{RMSE} \\
        \hline
        \cmark & \cmark & \xmark & \cmark & \textbf{0.96 ± 0.02} & \textbf{0.91 ± 0.03} & \textbf{2.58 ± 0.49} & \textbf{1.49 ± 0.28} \\
        \xmark & \cmark & \xmark & \cmark & 0.95 ± 0.03 & 0.91 ± 0.03 & 2.60 ± 0.50 & 1.50 ± 0.29 \\
        \cmark & \xmark & \xmark & \cmark & 0.95 ± 0.03 & 0.91 ± 0.03 & 2.61 ± 0.51 & 1.51 ± 0.30 \\
        \xmark & \xmark & \xmark & \cmark & 0.95 ± 0.03 & 0.91 ± 0.03 & 2.66 ± 0.55 & 1.53 ± 0.32 \\
        \xmark & \xmark & \cmark & \xmark & 0.93 ± 0.04 & 0.88 ± 0.04 & 3.04 ± 0.69 & 1.75 ± 0.40 \\
        \hline
    \end{tabular}}
    \label{tab:app-table1}
\end{table}

\begin{figure}[H]
    \centering
    \includegraphics[width=\textwidth]{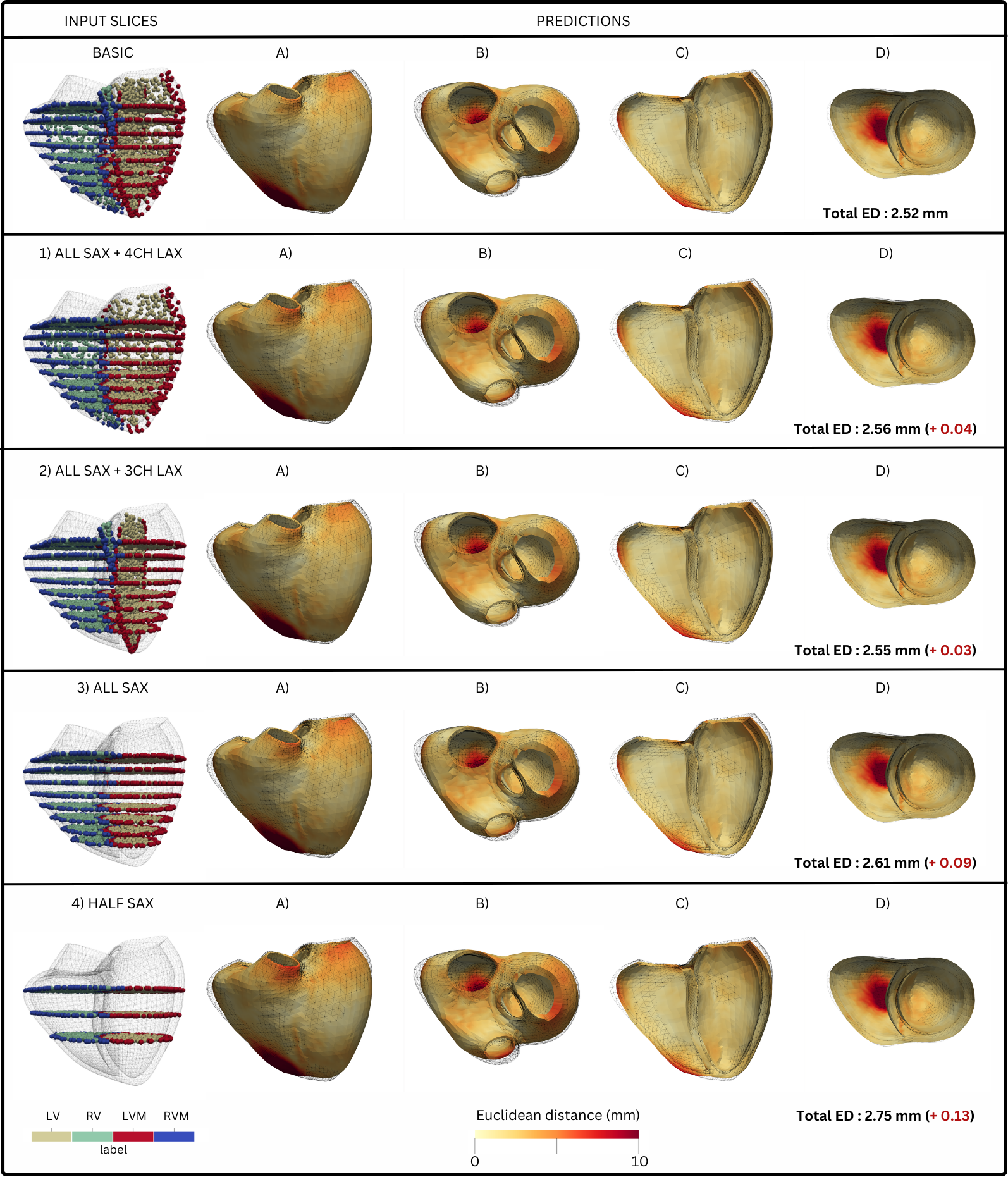}
    \caption[SYN ablation median]{Predictions of ablation study for the median case in the synthetic contour group. On the left, the input points used at inference optimisation in all the ablation cases are displayed. On the right, the predicted mesh is shown with a coloured colormap based on Euclidean distance from reference mesh, whereas the reference mesh is overlayed with a wireframe. The Euclidean distance from the reference mesh is shown with its relative difference to the full experiment (result in Fig.\ref{fig:res5}- median case). All shapes are equally scaled to preserve size information.}
    \label{fig:app-res3}
\end{figure}

\begin{figure}[H]
    \centering
    \includegraphics[width=\textwidth]{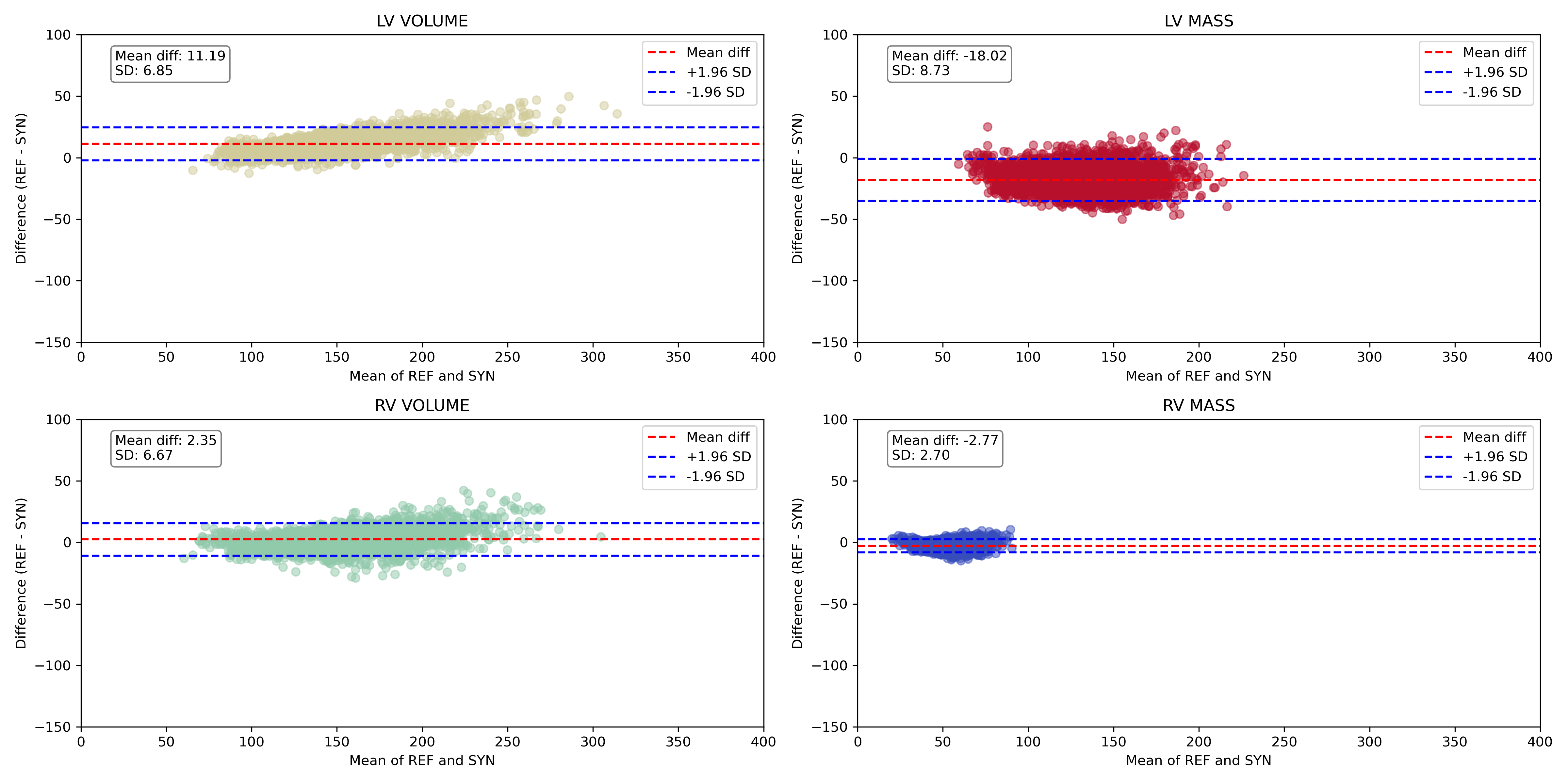}
    \caption[BA SYN CVD]{Bland-Altmann plots showing difference between volumes computed from REF meshes and predictions of SYN group for the CVD test set.}
    \label{fig:app-res4}
\end{figure}

\begin{figure}[H]
    \centering
    \includegraphics[width=\textwidth]{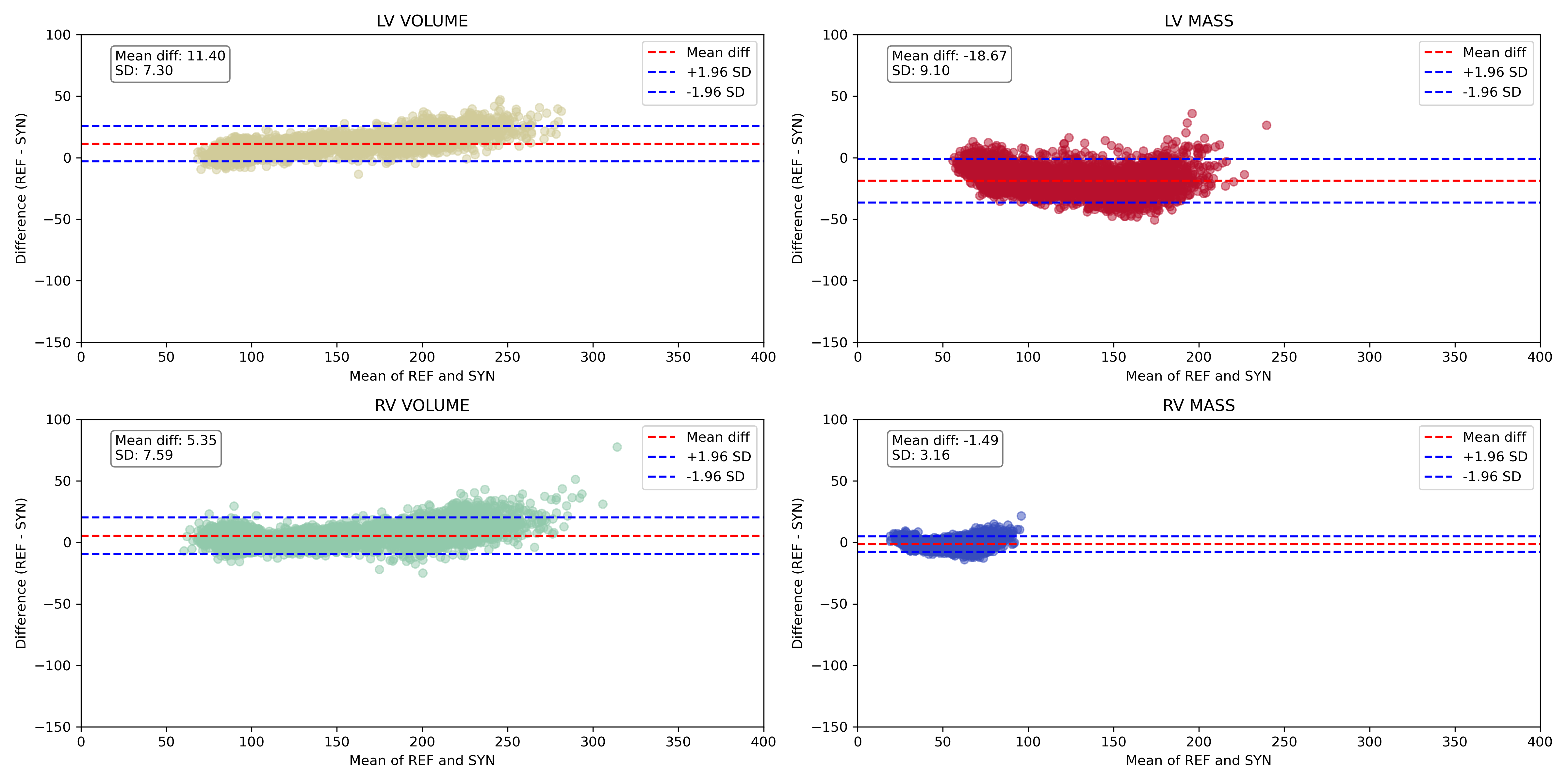}
    \caption[BA SYN No-CVD]{Bland-Altmann plots showing difference between volumes computed from REF meshes and predictions of SYN group for the No-CVD test set.}
    \label{fig:app-res5}
\end{figure}

\begin{figure}[H]
    \centering
    \includegraphics[width=\textwidth]{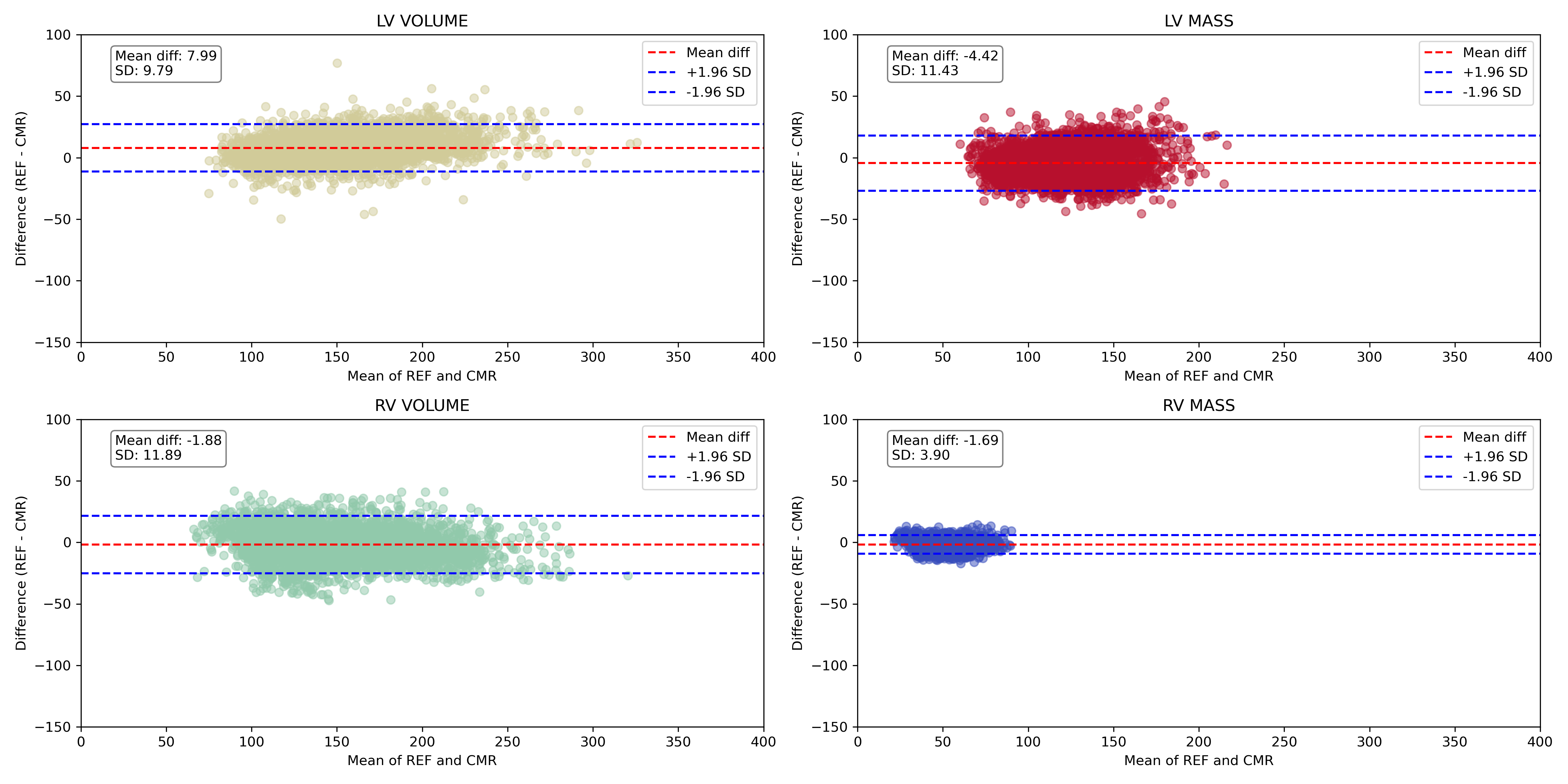}
    \caption[BA SEG CVD]{Bland-Altmann plots showing difference between volumes computed from REF meshes and predictions of SEG group for the CVD test set.}
    \label{fig:app-res6}
\end{figure}

\begin{figure}[H]
    \centering
    \includegraphics[width=\textwidth]{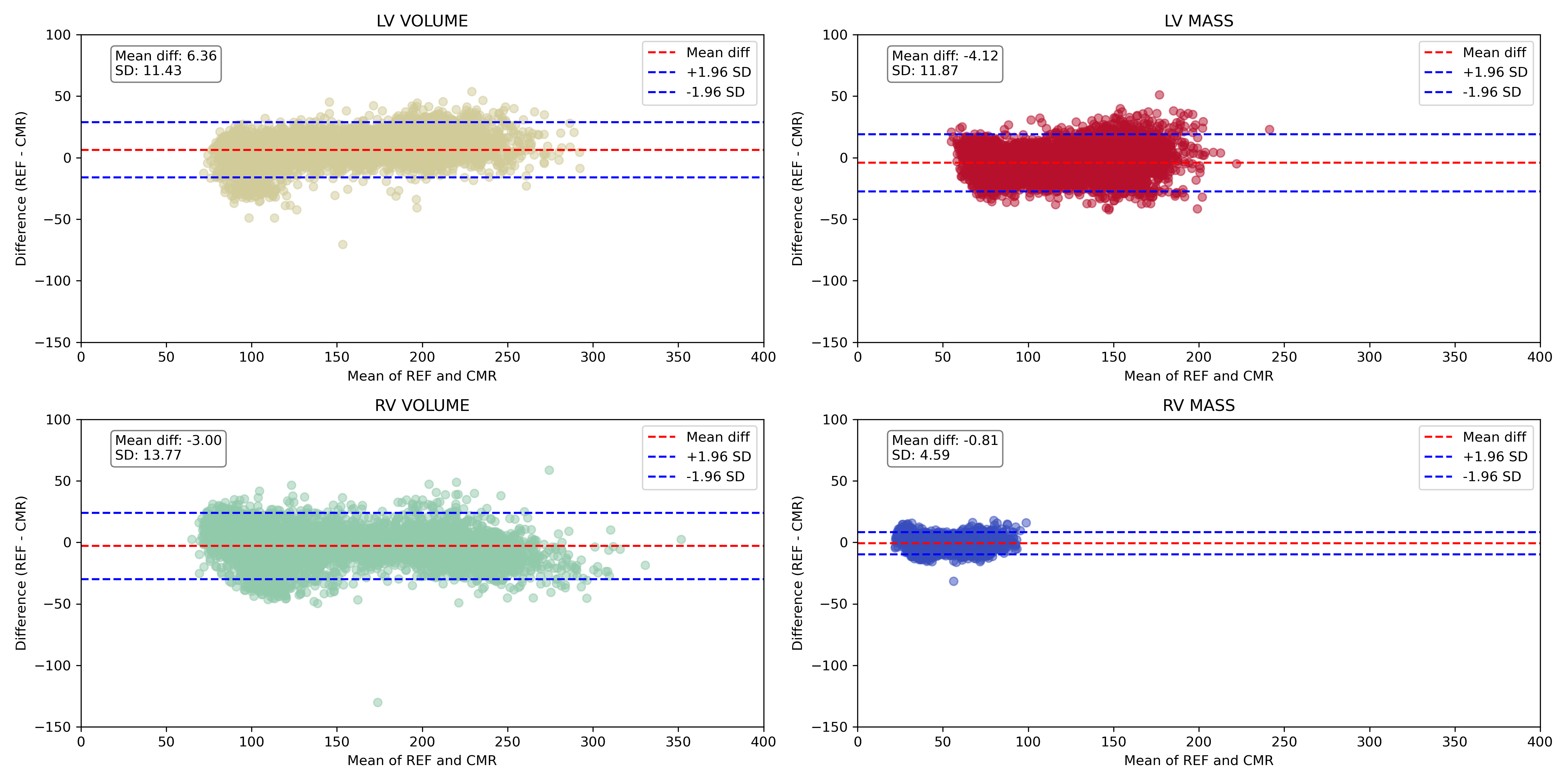}
    \caption[BA SEG No-CVD]{Bland-Altmann plots showing difference between volumes computed from REF meshes and predictions of SEG group for the No-CVD test set.}
    \label{fig:app-res7}
\end{figure}

\begin{figure}[H]
    \centering
    \includegraphics[width=\textwidth]{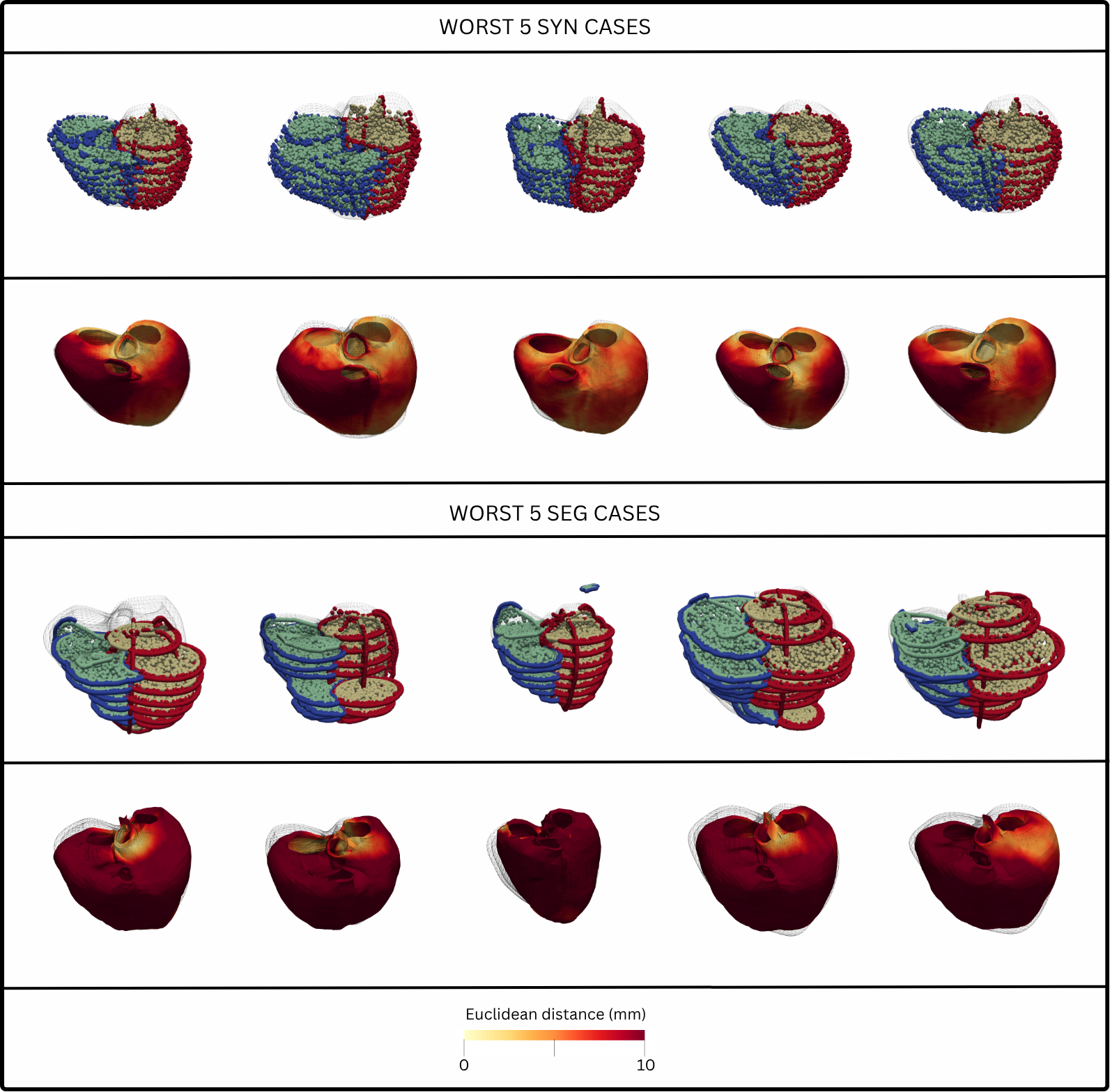}
    \caption[Pred worst  cases]{Predictions for the 5 worst SYN and SEG cases.}
    \label{fig:app-res8}
\end{figure}

\end{document}